\documentclass[journal]{IEEEtran}
\usepackage{amsmath}
\usepackage{amssymb}
\usepackage{booktabs}
\usepackage{graphicx}
\usepackage{subfigure}
\usepackage{float}
\usepackage{url}
\usepackage{multirow}
\usepackage{amsthm}
\usepackage{threeparttable}
\usepackage[linesnumbered,boxed,ruled,commentsnumbered,noend]{algorithm2e}
\usepackage[colorlinks,allcolors=black]{hyperref}
\SetKwIF{If}{ElseIf}{Else}{if}{}{else if}{else}{end if}%

\newtheorem{lemma}{Lemma}

\newtheorem{theorem}{Theorem}
\newtheorem{corollary}{Corollary}
\theoremstyle{definition}
\newtheorem{remark}{Remark}
\newtheorem{problem}{Problem}
\newtheorem{definition}{Definition}

\ifCLASSINFOpdf

\else

\fi

\usepackage{xcolor}

\begin{document}
%
\title{Deadlock Resolution  and Recursive Feasibility in MPC-based Multi-robot Trajectory Generation}
%
%
%

\author{
	Yuda Chen, Meng Guo and  Zhongkui Li 
	\thanks{This work was supported by the National Natural Science Foundation of China under grants U2241214, T2121002, 62373008, and 62203017. }
	\thanks{
		The authors are with the State Key Laboratory for Turbulence and
		Complex Systems, Department of Mechanics and Engineering Science,
		College of Engineering, Peking University, Beijing 100871, China.
Corresponding author: Zhongkui Li,
		\texttt{zhongkli@pku.edu.cn}.
	}
}

%
%

\maketitle

\begin{abstract}
	Online collision-free trajectory generation within a shared workspace is fundamental for most multi-robot applications.
	However, many widely-used methods based on model predictive control (MPC) lack theoretical guarantees on the feasibility of underlying optimization.
	Furthermore, when applied in a distributed manner without a central coordinator, deadlocks often occur where several robots block each other indefinitely.
	Whereas heuristic methods such as introducing random perturbations exist,  no profound analyses are given to validate these measures.
	Towards this end, we propose a systematic method called infinite-horizon model predictive control with deadlock resolution.
	The MPC is formulated as a convex optimization over the proposed modified buffered Voronoi with warning band.
    Based on this formulation, the condition of deadlocks is formally analyzed and proven to be analogous to a force equilibrium.
	A detection-resolution scheme is proposed, which can effectively detect deadlocks online before they even happen.
	 {
		Once detected, it utilizes an adaptive force scheme to resolve deadlocks,
		under which no stable deadlocks can exist under minor conditions on robots' target positions.
	}
	In addition, the proposed planning algorithm ensures recursive feasibility of the underlying optimization at each replanning under both input and model constraints, is concurrent for all robots and requires only local communication.
	Comprehensive simulation and experiment studies are conducted over large-scale multi-robot systems.
	Significant improvements on success rate are reported, in comparison with other state-of-the-art methods
        and especially in crowded and high-speed scenarios.
\end{abstract}


%
\IEEEpeerreviewmaketitle

\section{Introduction}\label{section:introduction}
%
%
%
%
\IEEEPARstart{C}{ollision}-free
trajectory generation is essential for multi-robot systems to perform various missions in a shared environment, such as cooperative inspection and transportation \cite{Chung2018,Zhou2022,Zhang2021,Gafur2021}.
However, it becomes especially challenging when a large number of agile robots navigate in a crowded space with high speed.
The commonly-seen multi-robot trajectory generation (MATG) algorithms can be classified into roughly six categories:
potential fields~\cite{Borenstein1989,Koren1991} that design virtual driving forces induced by artificial potentials;
 {
geometric guidance~\cite{Alonso2012,van2008} such as reciprocal velocity obstacles (RVO) \cite{van2008,van2011,Huang2023} that analyze the geometric properties based on the position and velocity of the others;
}
conflicts resolution \cite{Tomlin1998,Cole2018} that designs heuristic rules to resolve potential collisions;
learning-based methods \cite{Fan2020,Shi2021,Chen2017} that rely heavily on accurate simulators and reward shaping;
control-law-based methods~\cite{Zhao2018,Panagou2016,Panagou2020} that have a strong theoretical guarantee;
and optimization-based methods \cite{Richards2004,Ferik2016,Cichella2021} that model the problem as numerical optimizations.

Optimization-based methods have gained increasing popularity recently due to its modeling capability and extensibility.
In particular, optimization-based methods construct and solve various optimizations to achieve collision-free navigation, such as the mixed integer quadratic programming in \cite{Richards2004}, sequential convex programming in \cite{Augugliaro2012}, and model predictive control (MPC) in \cite{Morgan2016}.
However, most of the aforementioned methods do not ensure explicitly feasibility of the underlying optimization during the whole navigation.
In other words, the optimization might be infeasible at one time step, thus the whole system stops.
Some works in \cite{Park2021}, \cite{Honig2018} and \cite{Leobardo2017} propose to tackle the feasibility problem by gradually scaling up the time step, which however requires a centralized coordinator.
Another  work \cite{Wang2017} introduces the notion of control barrier function, which can guarantee collision-free trajectory.
Nonetheless, it  might be overly conservative due to excessive breaking.
The methods in~\cite{Park2022} ensure feasibility by utilizing relative-safe corridor among robots.
Last but not least, the above methods often impose a fully-connected communication network with high communication burden.

More importantly, another well-known issue in MATG is that deadlocks often occur during navigation in multi-robot systems.
Formally, a deadlock happens when multiple robots are blocked by each other indefinitely and cannot make any progress towards their targets~\cite{Alonso2018-2}.
It is often caused by the symmetric configuration of the underlying system \cite{Grover2020}, and the lack of a central coordinator~\cite{Alonso2018-1}.
A commonly-used resolution scheme is inspired by the right-hand rules.
For instance, an artificial perturbation to the right-hand side of each robot is introduced in \cite{Wang2017} in order
to break the equilibrium of deadlock.
This however can lead to unpredictable behavior and even safety issue in practice as the magnitude of such perturbations is hard to determine.
Other works in~\cite{Zhou2017,Pierson2020,Abdullhak2021} instead propose to select a detour point on the right-hand side of each robot as the temporary target.
Nonetheless, the validity and effectiveness of these schemes still lack theoretical analyses and guarantee on the performance, and might lead to livelock problems in practice where the robots oscillate around the deadlock positions indefinitely.
Similar clockwise maneuver is adopted in~\cite{Garg2019} which shows that the tangent motion can resolve deadlocks for some cases.
Important advances on deadlock analysis and resolution are recently reported in~\cite{Grover2020,Grover2020-1,Grover2022}. The authors in~\cite{Grover2020,Grover2020-1,Grover2022} analyze the condition of deadlocks based on the Karush-Kuhn-Tucker (KKT) formulation for multi-robot systems and present a proportional-derivative control law to resolve deadlock, which can be theoretically shown for the case with no more than three robots.
As pointed out in \cite{Zhou2017}, algorithms that can provably avoid deadlock in general cases without a global coordinator still await.

To address the open problem of deadlock resolution with feasibility guarantee in a distributed manner, this work proposes a novel systematic trajectory generation method, called infinite horizon model predictive control with deadlock resolution (IMPC-DR).
Firstly, the traditional buffered Voronoi cells (BVC) proposed in \cite{Zhou2017} is extended by taking into account full planned trajectories of neighboring robots and introducing a velocity-dependent buffer width.
Furthermore, an extra warning band is added to the terminal horizon to facilitate the resolution of potential deadlocks.
An improved distributed MPC formulation is proposed based on these constraints and a novel cost function that deals with potential deadlocks.
Given this formulation, the condition of deadlocks during navigation is formally analyzed and shown to be analogous to a force equilibrium between the attractive forces from the targets and repulsive forces from the neighboring robots.
Consequently, a detection mechanism is designed to detect any potential deadlocks online and before they might happen.
Once detected, an adaptive resolution scheme is followed to incrementally adjust the inter-robot repulsive force to break the equilibrium.
Properties  of the proposed algorithm regarding deadlock resolution, recursive feasibility and local communication are all formally analyzed and proven.

To summarize, the main contributions of this paper are as follows:

\begin{itemize}

\item[$\bullet$] The modified buffered Voronoi cells with warning band (MBVC-WB) introduces a novel trajectory-based and velocity-dependent space partition technique.  
It can avoid  potential collision between the consecutive sampling points, which may take place in the traditional point-based BVC in~\cite{Zhou2017} and the sampling-based methods in, e.g., ~\cite{Morgan2016,Zhu2019}.

\item[$\bullet$] The condition of deadlocks for MPC-based MATG problems is formally analyzed and revealed as a force equilibrium.
This is more general than an earlier result in \cite{Grover2020} for three robots.
Based on this condition, a novel online detection scheme is provided to detect potential deadlocks early-on before they appear, i.e., not afterwards as in~\cite{Wang2017,Zhou2017}.
Furthermore, in contrast to heuristic methods in~\cite{Wang2017,Pierson2020, Abdullhak2021},
the proposed deadlock resolution scheme has a theoretical demonstration that the inter-robot repulsive forces are adapted in a smooth way to falsify the deadlock condition, again before potential deadlocks appear. 
It is formally proven that under the proposed resolution scheme, no stable deadlocks can exist under minor conditions on the target positions.

\item[$\bullet$] The proposed complete algorithm not only ensures the deadlock resolution, but also guarantees that the optimization at each replanning is recursively feasible. 
Such an assurance for feasibility is often overlooked and simply assumed in related works \cite{Augugliaro2012,Chen2015,Luis2019}, rather than guaranteed explicitly and formally.

\item[$\bullet$] Effectiveness and performance of the proposed algorithm are validated extensively by numerical simulations against other state-of-the-art methods including iSCP~\cite{Chen2015}, DMPC~\cite{Luis2019} and BVC~\cite{Zhou2017}.
Our method shows a significant increase in both success rate and feasibility, especially for large-scale crowed and high-speed scenarios.

\item[$\bullet$] Hardware experiments are carried out using nano quadrotors Crazyflies in a workspace captured by an indoor motion capture system OptiTrack.
Up to $20$ nano quadrotors successfully perform experiments of antipodal transitions in 2D and 3D,
validating the applicability of the proposed algorithms on real-time platforms.

\end{itemize}

As mentioned earlier, the problem of deadlock resolution considered in this paper is restricted to the MATG area.  {It should be mentioned that deadlock phenomena also take place in other areas such as computer sciences~\cite{Zhou2005}} and discrete event systems \cite{Jonghun2001,Reveliotis2022,Elzbieta2013,Zhou2020,Xing2022}. 
For instance, the works~\cite{Jonghun2001,Reveliotis2022,Elzbieta2013} address deadlock resolution in multi-robot path planning on topological graphs in the area of traffic management.
Although the deadlock phenomena in different areas are similar, yet the root cause, the resolution schemes and the implemented tools are quite distinct.
More specifically, the deadlocks considered here belongs to the category of policy-induced deadlocks, due to the distributed and online trajectory planning scheme.
In contrast, the deadlocks from the aforementioned works are often caused by the topological structure and the transition condition of the robots.
Thus, the associated resolution schemes and the involved tools are rather distinct. 
The former requires the real-time adaptation of robot trajectories under dynamical and geometric constraints,
while the latter often focuses on designing discrete policies using tools such as finite-state automata or labeled transition systems.
  

The remaining parts of this paper are organized as follows.
Section~\ref{section:problem-statement} describes the problem statement.
The method and the corresponding analysis are illustrated in Section~\ref{section:technical-method} and Section~\ref{section:analysis}, respectively.
Section~\ref{section:simulation-and-experiment} includes numerical simulations and hardware experiments.
Conclusions and future work can be found in Section~\ref{section:conclusion}.

\section{Problem Statement} \label{section:problem-statement}

This section formulates the multi-robot trajectory generation (MATG) problem considered in this paper.

\subsection{ Robot Dynamics }

Consider a team of~$N$ robots, where each robot~$i \in \mathcal{N} \triangleq \{1,2,\cdots,N\}$ is modeled as a point mass in~$\mathbb{R}^d$, and~$d={2,3}$ is the dimension of the configuration space.
Furthermore, its motion is approximated by the double integrator, i.e.,
\begin{equation*}
	x^i(t+h)=\mathbf{A} x^i(t) + \mathbf{B} u^i(t),
\end{equation*}
where $x^i(t)=\left[{p^i}(t),{v^i}(t)\right]$ is robot $i$'s state, including the position~$p^i(t)$ and velocity~$v^i(t)$,   $u^i(t)$ as acceleration is the control input, 
$\mathbf{A}
=
\left[
\begin{smallmatrix}
	\mathbf{I}_d & h \mathbf{I}_d  \\
	\mathbf{0}_d & \mathbf{I}_d \\
\end{smallmatrix}
\right] ,\,\mathbf{B}=
\left[ \begin{smallmatrix}
	\mathbf{0}_d   \\
	h \mathbf{I}_d
\end{smallmatrix}
\right]$
and $h$ is the sampling interval. For brevity, the time index~``$(t)$" is omitted wherever ambiguity is not caused in the sequel.
As required in practice, both the robot's velocity and acceleration are subjected to physical constraints. Specifically, it holds that~$\| \Theta_v v^i \| \le~v_{\text{max}}$  and~$\| \Theta_a u^i \| \le a_{\text{max}}$, where~$\Theta_v, \Theta_a$ are positive-definite matrices, and $v_{\text{max}},\, a_{\text{max}}>0$ denote the maximum velocity and acceleration, respectively.

\subsection{Collision Avoidance}

To avoid inter-robot collisions, the minimum distance allowed between any pair of robots is set to~$r_{\text{min}}>0$.
In other words, collision is avoided, if
\begin{equation} \label{collision}
	\| p^{ij} \|= \| p^i-p^j \| \ge r_{\text{min}}
\end{equation}
holds, for \emph{any} pair $(i,\,j)\in \mathcal{N}\times \mathcal{N}$ and $i\neq j$.

\subsection{MPC-based MATG} \label{subsection:MATG}

The general MATG problem is to design control inputs~$u^i$ for each robot~$i\in \mathcal{N}$ such that it reaches the target position~$p^i_{\text{target}}\in \mathbb{R}^d$, while avoiding collisions with other robots.
As discussed in Section~\ref{section:introduction}, many existing work adopts the MPC-based solutions, e.g.,~\cite{Richards2004,Morgan2016}, where the robot trajectories are calculated by solving an optimization problem at each time step and then executed in a receding horizon fashion.
At each time step~$t\geq t_{0}$, the planned trajectory of robot $i$ for the future~$K$ time steps is defined as
$\mathcal{P}^i=[p^i_1,\,p^i_2,\,\ldots,\,p^i_K]$
where $p^i_k$ is the planned position for time instant~$t + k h$, $k\in \mathcal{K}$;
$\mathcal{K} \triangleq \{1,2,\cdots,K\}$ and $K$ is the length of the planning horizon.
As a rule of thumb,
$K > \frac{v_\text{max}}{a_\text{max}h}$ is chosen such that
the robot can reduce its velocity from maximum to zero within the horizon.
Similar notations apply to~$v^i_k(t)$, $u^i_k(t)$ as the planned velocity and acceleration.
For the ease of notation, $\boldsymbol{u}^i= [u^i_0,u^i_1,\ldots,u^i_{K-1}]$ and
analogously for~$\boldsymbol{x}^i$.
Then, the following constrained optimization problem is imposed for the MATG at time~$t>t_0$.

\begin{problem} \label{problem}
	At time~$t>t_0$,
	the planned trajectory~$\mathcal{P}^i$ of robot~$i\in \mathcal{N}$ is the solution to
        the following optimization:
	\begin{subequations}\label{eq:original-mpc}
		\begin{align}
			\min _{\{\boldsymbol{u}^{i},\boldsymbol{x}^i\}}
			& \  C^i \left( \boldsymbol{u}^{i},\boldsymbol{x}^i \right)  \label{eq:original-mpc-cost}\\
			\textbf{s.t.}\quad
			& {\| p^i_k-p^j_k  \|} \ge r_{\text{min}}, \;\forall j \ne i, \forall k  ;\label{eq:original-mpc-collision} \\
			& x_{k}^{i}=\mathbf{A} x_{k-1}^{i}+\mathbf{B} u_{k-1}^{i}, \;\forall k ; \label{eq:dynamic-constraint}\\
			&\| \Theta_a u^i_{k-1} \| \le a_{\text{max}},\;\forall k  ; \label{eq:input-constraint}\\
			& \| \Theta_v v_k^i \| \le v_{\text{max}}, \;\forall k  ; \label{eq:velocity-constraint}
		\end{align}
	\end{subequations}
	where $\forall i,\,k$ is an abbreviation for~$\forall i\in \mathcal{N}$ and~$\forall k\in \mathcal{K}$;
    $C^i(\cdot)$ is the cost function to be minimized.
	\hfill $\blacksquare$
\end{problem}

\begin{remark}
  The objective function in~\eqref{eq:original-mpc-cost} can be of various formats, e.g.,
  $C^i = \frac{1}{2} \sum_{k}(  Q_k \| p_{k}^{i}-p^i_{\text{target}} \|^2 + R_k  \| u_{k}^{i} \|^2 )$
  in~\cite{Zhou2017}, where $Q_k,R_k > 0$, $k \in \mathcal{K}$.
  As discussed in the sequel, a novel objective formulation is proposed in this work,
  which incorporates the deadlock resolution as an integral part of the optimization process.
  \hfill $\blacksquare$
\end{remark}

Once this optimization is solved at time~$t$ and the planned trajectory is derived, the low-level feedback controller of each robot~$i$ tracks~$\mathcal{P}^i(t)$ during the time interval~$\left[ t,\, t+h \right)$.
Consequently, it holds that~$x^i(t+h)=x^i_1(t)$ at time~$t+h$.
Afterwards, the above optimization is re-formulated given the updated system state, and solved again for the planned trajectory~$\mathcal{P}^i(t+h)$.
Each robot repeats this procedure locally until \emph{all} robots reach their respective target positions.
This is the common framework for most MPC-based MATG methods.

\subsection{Distributed Solution, Recursive Feasibility and Deadlock}

Notably, constraint~\eqref{eq:original-mpc-collision} depends on the state of robot~$j$,
meaning that the optimization in~\eqref{eq:original-mpc} cannot be solved by robot~$i$ alone.
However, a \emph{distributed} solution without a central coordinator is desired in this work.
In other words, the planned trajectories of all robots are calculated parallel and locally by each robot solving the above optimization.
Although each robot can only dictate its own trajectory, it can exchange data with other robots via wireless communication, which is assumed to be bidirectional and ideal without drop-outs or delays.
Furthermore, as mentioned previously in Section~\ref{section:introduction}, there are two aspects of Problem~\ref{problem} that are of particular interest in this work: 
recursive feasibility and deadlock.

According to \cite{Rawlings2017}, the recursive feasibility of the MPC-based MATG formulated in Problem 1 is defined as follows. 

\begin{definition}[Recursive Feasibility]\label{def:feasibility} 
	 { 
		The constrained optimization \eqref{eq:original-mpc} is called recursive feasible, if it is feasible at time step~$t$, $\forall t>t_0$, for each robot~$i\in \mathcal{N}$, then the new optimization at time step~$t+h$ is feasible as well.
	}
	\hfill $\blacksquare$
\end{definition}

Recursive feasibility ensures the safety of the resulting trajectories, namely, no inter-robot collisions would happen. However, for certain cases such as symmetric and crowded scenarios, a number of robots block each other and cannot move towards their target positions, also known as deadlocks~\cite{Wang2017,Alonso2018-2,Grover2022}. 
Although no collisions happen, this still prohibits a successful navigation.
 {
	For the MATG problem considered here, the deadlock is formally defined as follows.
}

\begin{definition}[Deadlock]\label{def:deadlock}
	 {
		Robot~$i$ is said to be in a deadlock, if its planned positions $p^i_k$ remain static
		at its current position $p^i$ and it does not reach its target, i.e., 
		$p^i(t) = p^i_k(t)=p^i(t_{\text{d}}) \neq p_\text{target}^i$, $\forall k\in \mathcal{K}$, $\forall t>t_{\text{d}}$, where~$t_{\text{d}}$ is the time when this deadlock starts.
	}
	\hfill $\blacksquare$
\end{definition}

 {
	The main objective of this paper is to design an MPC-based trajectory planning method such that i) the  constrained optimization~\eqref{eq:original-mpc}  is recursive feasible; ii) the underlying robots can detect and resolve potential deadlocks.
}

\begin{table} [t!]
	\caption{Nomenclatures}
	\begin{tabular}{ll}
		\toprule
			$p^i_\text{target}$ & Target of robot $i$.\\
			$p^i$, $p^{ij}$      & Position of robot~$i$, $p^{ij}=p^i-p^j$.\\
			$p^i_k$  & Planned position of robot $i$ at the $k$-th step of horizon.\\
			$\mathcal{P}^i$ & Planned trajectory of robot $i$.\\
			$\overline{\mathcal{P}}^i$ & Predetermined trajectory of robot $i$.\\
			$\overline{p}^i_k$ & Position of robot~$i$ at the $k$-th step of horizon in $\overline{\mathcal{P}}^i$.\\
			$v^i_k$    & Velocity of robot $i$ at the $k$-th step of horizon.\\
			$x^i_k$    & $x^i_k=[p^i_k,v^i_k]$.\\
			$u^i_k$    & Control input of robot~$i$ at the $k$-th step of horizon.\\
			$K$        & Length of horizon.\\
			$\mathcal{K}$, $\tilde{\mathcal{K}}$ & $\mathcal{K} = \left\{ 1,2,\ldots,K \right\}$ and $\tilde{\mathcal{K}} = \left\{ 1,2,\ldots,K-1 \right\}$.\\
			$w^{ij}$   & Warning band of robot $i$ w.r.t. robot $j$.\\
			$r_\text{min}$ & The minimum allowable distance between any two robots.\\
			$r^\prime_\text{min}$ & $r^\prime_\text{min} = \sqrt{r_{\text{min} }^{2}+h^{2} v_{\text{max} }^{2}}$.\\
			$\mathcal{N}$, $\mathcal{N}^i$ & The set of all robots, and 
			$\mathcal{N}^i = \left\{ j \ | w^{i j} < \epsilon \right\}$\\
			$\rho^{ij}$ & The repulsive coefficient of robot $i$ w.r.t. robot~$j$.\\
			$C^i$     & The objective function of robot $i$.\\
		\bottomrule
	\end{tabular}
\end{table}

\section{Proposed Solution} \label{section:technical-method}

The complete solution is described in this section,
which consists of four main steps:
re-formulation of the constraint in~\eqref{eq:original-mpc-collision} for collision avoidance;
re-formulation of the complete optimization;
formal analyses of the condition for deadlocks;
and finally the resolution scheme of potential deadlocks.

\subsection{MBVC-WB} \label{subsection MBVC}

The collision avoidance constraint in~\eqref{eq:original-mpc-collision} is enforced explicitly by requiring the inter-robot distance to be more than~$r_{\min}$ at all time.
However, the future states of other robots are not available at the current time. 
Thus, we propose to replace them with the \emph{predetermined  trajectory} of other robots.

\begin{figure}[t!]
	\centering
	\includegraphics[width=1.0\linewidth]{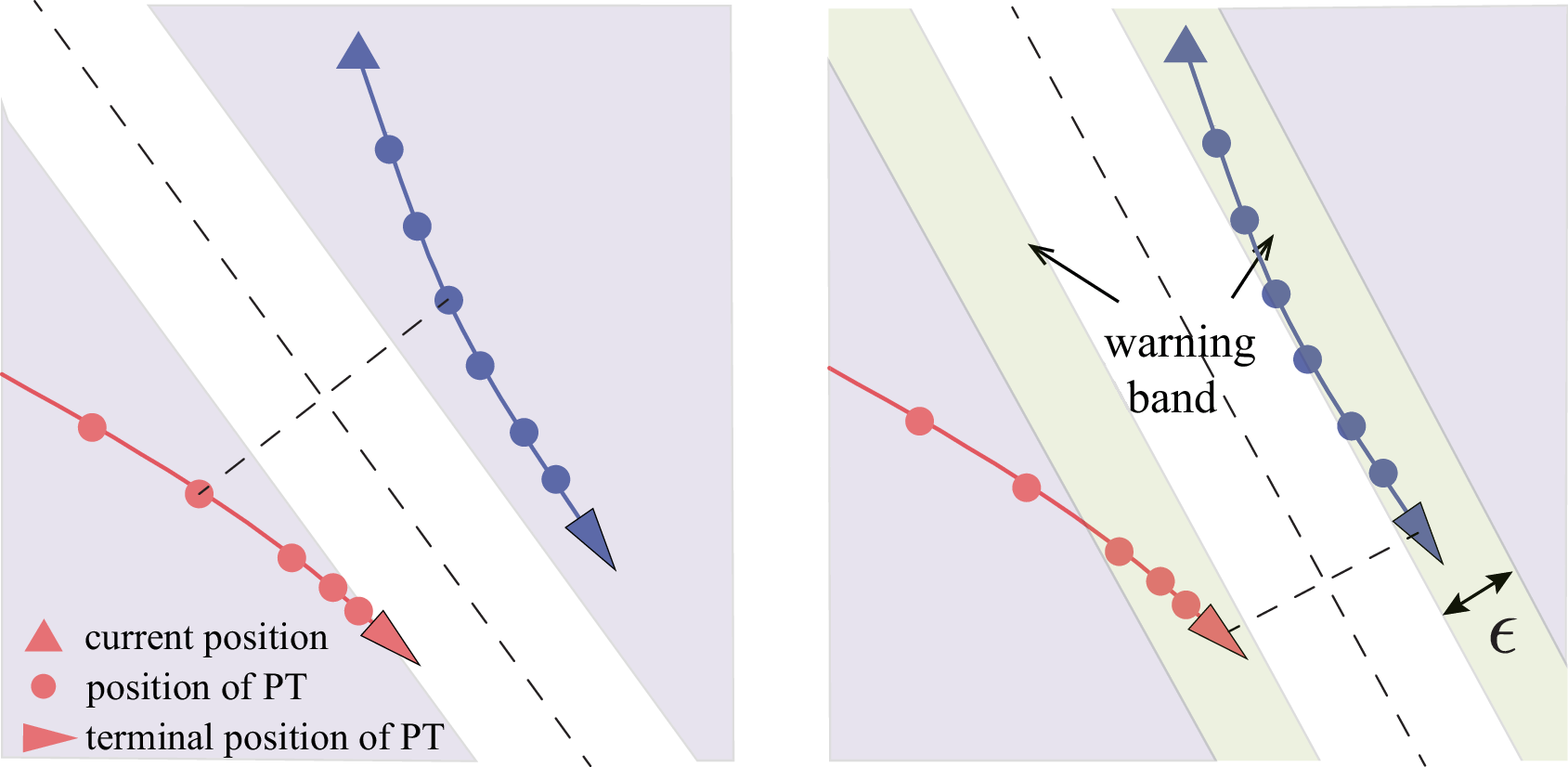}
	\caption{ 
		Illustration of the MBVC-WB.  
		\textbf{Left}: the shared space is split at each step of horizon;
  		\textbf{Right}: a warning band is added for the terminal step of horizon, i.e., $k=K$.
  	}
	\label{MBVC-WB}
\end{figure}

\begin{definition}[Predetermined Trajectory (PT)]\label{def:pt}
	The predetermined trajectory for robot $i$ at time $t$ is defined as
	\begin{equation*}
		\overline{\mathcal{P}}^i(t)=\left[ \overline{p}_{1}^{i}(t), \overline{p}_{2}^{i}(t), \ldots, \overline{p}_{K}^{i}(t) \right],
	\end{equation*}
	where $ \overline{p}_{k}^{i}(t) = p^i_{k+1}(t-h),  k\in \tilde{\mathcal{K}} \triangleq \{1,\cdots, K-1\}$, and $\overline{p}^i_{K}(t)=p^i_{K}(t-h)$ for the end of horizon.
	\hfill $\blacksquare$
\end{definition}

Based on the predetermined trajectory, we use the spatial separation method as in~\cite{Park2022} to handle inter-robot collisions.
As illustrated in Fig.~\ref{MBVC-WB}, this method forms a separating hyperplane between different robots which restricts their respective motion space.
Inspired by~\cite{Zhou2017}, we define the following modified buffered Voronoi cell with warning band (MBVC-WB) for any pair of robots $i$ and $j$ that $j \neq i$:
\begin{equation} \label{eq: MBVC definition}
	\begin{split}
		\mathcal{V}^{ij}_k=\left\{ p  \, |\,    (p-\frac{\overline{p}^i_k+\overline{p}^j_k}{2})^{\mathrm{T}}
		\frac{ \overline{p}_{k}^{ij} }
		{ \|\overline{p}_{k}^{ij}\| }
		 \ge r_k^{ij} \right\},
	\end{split}
\end{equation}
where~$\mathcal{V}^{ij}_k$ is the MBVC-WB for robot~$i$ w.r.t. robot~$j$ over the horizon~$k\in \mathcal{K}$;
$\overline{p}_{k}^{ij}=\overline{p}_{k}^{i}-\overline{p}_{k}^{j}$;
the pair-wise buffer width~$r_k^{ij}=\frac{ r^{\prime}_{\text{min}} }{2}$ holds for~$k\in \tilde{\mathcal{K}}$, where
\begin{equation}\label{eq:r-prime}
	r^{\prime}_{\text{min}} = \sqrt{r_{\text{min} }^{2}+h^{2} v_{\text{max} }^{2}}
\end{equation}
is the extended buffer width;
and~$r_K^{ij}=\frac{ r^{\prime}_{\text{min}} }{2}+w^{i j}$,
where $w^{i j}$ is an optimization variable as the warning distance between robots~$i$ and~$j$.
Note that $0 \leq w^{i j} \leq \epsilon$ with $\epsilon$ being the maximum width of the warning band, which is generally chosen as $\epsilon\in (\frac{r_\text{min}}{6},\frac{r_\text{min}}{2})$.

Like BVC in \cite{Zhou2017}, an important role of MBVC-WB is that it can convert the original non-convex constraint~\eqref{eq:original-mpc-collision} into two convex constraints.
It is not difficult to derive that the collision avoidance constraints between any pair of robots can be decoupled and become~$p^i_k \in \mathcal{V}^{i j}_{k}$, $p^j_k \in \mathcal{V}^{j i}_{k}$.
Via simple re-arrangements, the constraint $p^i_k \in \mathcal{V}^{i j}_{k}$ can be rewritten as
\begin{subequations} \label{eq: a p > b}
	\begin{align}
		&{a_{k}^{i j}}^\mathrm{T} p_{k}^{i} \geq b_{k}^{i j}, \ \forall j\neq i, k\in \tilde{\mathcal{K}}, \\
		&{a_{K}^{i j}}^\mathrm{T} p_{K}^{i} \geq b_{K}^{i j} + w^{i j}, \forall j \neq i;
	\end{align}
\end{subequations}
where the coefficients are given by
\begin{equation} \label{a-b-definition}
	a_{k}^{i j}=\frac{ \overline{p}_{k}^{ij} } { \|\overline{p}_{k}^{ij}\| },\quad
	b_{k}^{i j}={a_{k}^{i j}}^\mathrm{T} \frac{\overline{p}_{k}^{i} + \overline{p}_{k}^{j}}{2}+\frac{r_{\min }^{\prime}}{2}.
\end{equation}
Note that the above constraints in~\eqref{eq: a p > b} can be formulated locally by each robot given the planned trajectory~$\mathcal{P}^j(t-h)$ received from its neighbors~$j$ at the previous time step~$t-h$.

\begin{remark}
The buffered Voronoi cell (BVC) proposed in \cite{Zhou2017} is defined as follows:
\begin{equation*}
		\overline{\mathcal{V}}^{ij}=\left \{ p \, | \, (p-\frac{p^i+p^j}{2}) ^{\mathrm{T}}
		\frac{ p^{ij} }
		{ \| p^{ij} \| }
	\ge \frac{1}{2}r_{\text{min}} \right \}.
\end{equation*}
Compared with the proposed MBVC-WB in~\eqref{eq: MBVC definition}, there are three main differences:
(i) BVC only considers the \emph{current} positions of robots~$i$ and~$j$, while MBVC-WB takes into account all future positions of both robots according to their predetermined trajectories.
This leads to a more accurate space separation and thus a higher utilization rate of the workspace;
(ii) the buffer width~$r_{\min }^{\prime}$ in~\eqref{eq:r-prime} is velocity
dependent to ensure collision avoidance between sampling points;
(iii) as explained in the sequel, the warning band~$w^{ij}$ added at the end of the horizon is also included in the objective function, which allows for {an effective resolution} scheme for potential deadlocks.
\hfill $\blacksquare$
\end{remark}

\begin{lemma} \label{lemma:MBVC}
	If $p^i_k \in \mathcal{V}^{i j}_{k}$ holds, $\forall i,j\in \mathcal{N}$ that~$i\neq j$,
	and $\forall k\in \mathcal{K}$, then
	\begin{equation} \label{MBVC property}
		\|p^{ij}_k\| \geq  r_{\min}^{\prime}
	\end{equation}
	holds for the same set of~$i,j,k$ as above and $p^{ij}_k = {p^i_k-p^j_k}$.
	Furthermore, assuming that robots~$i,\,j$ move at constant velocities from~$p^i_{k}$ to~$p^i_{k+1}$ and from~$p^j_{k}$ to~$p^j_{k+1}$, respectively, the planned trajectories~$\mathcal{P}^i(t),\, \mathcal{P}^j(t)$ are collision-free over the whole trajectory.
\end{lemma}

\begin{proof}
	For robots $i$ and $j$, if $p^i_k \in \mathcal{V}^{i j}_k$ and $p^j_k \in \mathcal{V}^{j i}_k$ hold for $\forall k \in \mathcal{K}$, it follows that ${a_{k}^{i j}}^\mathrm{T} p_{k}^{j} \geq b_{k}^{i j}$ and ${a_{k}^{j i}}^\mathrm{T} p_{k}^{j} \geq b_{k}^{j i}$, respectively. 
	Hence, ${a_{k}^{i j}}^\mathrm{T} p_{k}^{j} + {a_{k}^{j i}}^\mathrm{T} p_{k}^{j} \geq b_{k}^{i j} + b_{k}^{j i}$ holds.
	Substituting \eqref{a-b-definition} into it, it follows that ${a_{k}^{i j}}^\mathrm{T} p^{ij}_k \geq r^{\prime}_\text{min}$.
	Moreover, since~${a_{k}^{i j}}^\mathrm{T} p^{ij}_k \leq \| a_{k}^{i j} \|
	\| p^{ij}_k \| \leq \| p^{ij}_k \|$, \eqref{MBVC property} can be derived.
        Furthermore, due to~\eqref{MBVC property}, it can be shown that
	\begin{equation*}
		\|p^{ij}_{k}\| \geq r^{\prime}_{\text{min}} = \sqrt{r_{\text{min}}^{2}+h^{2} v_{\max }^{2}}
	\end{equation*}
    holds for both~$k$ and~$k-1$.
	Since the maximum allowed velocity is $v_{\text{max}}$, it follows that $\| p^{i}_{k}-p^{i}_{k-1} \|  \leq h \ v_{\max } $ and $\| p^{j}_{k}-p^{j}_{k-1} \| \leq h \ v_{\max }$.
	Consequently, it yields
	\begin{equation*}
		\begin{aligned}
			&\|p^{ij}_k\| \geq \sqrt{r_{\text{min} }^{2}+h^{2} v_{\text{max} }^{2}} \\
			& \geq \sqrt{r_{\text{min} }^{2}+\frac{1}{4}\|p^{i}_{k}-p^{i}_{k-1}-p^{j}_{k}+p^{j}_{k-1}\|^{2}}.
		\end{aligned}
	\end{equation*}
	Similarly, it holds that
	\begin{equation*}
		\begin{aligned}
			&\|p^{ij}_{k-1}\| \geq
			&\sqrt{r_{\min }^{2}+\frac{1}{4}\|p^{ij}_{k}-p^{ij}_{k-1}\|^{2}}.
		\end{aligned}
	\end{equation*}
	Given these conditions, Appendix~\ref{appendix:lemma-distance} shows that
        the inter-robot distance during time~$[t+(k-1)h,\,t+kh]$ is bounded as
	\begin{equation} \label{eq: interval minimum distance}
		\|p^i_{k-1}+\beta \left(p^i_k-p^i_{k-1}\right)-p^j_{k-1} - \beta \left(p^j_k-p^j_{k-1}\right)\| \geq r_{\min }
	\end{equation}
    where~$\beta \in [0,1]$.
    In other words, if robot~$i$ moves from $p^{i}_{k-1}$ to $p^{i}_{k}$ and robot~$j$ from $p^{j}_{k-1}$ to $p^{j}_{k}$ at constant velocities, according to~\eqref{eq: interval minimum distance}, the minimum distance between robots $i$ and $j$ is larger than $r_{\text{min}}$ for all~$k\in \mathcal{K}$.
	This completes the proof.
\end{proof}

\subsection{Complete Optimization}

In addition to the inter-robot collision avoidance,
the following terminal constraint is introduced to ensure the feasibility of optimization~\eqref{eq:original-mpc}:
\begin{equation} \label{static constraint}
	x_{K}^{i} \in X_{e},\; \text{where}\; X_{e}=\{x \mid x = \textbf{A} x + \textbf{B} u, u \in \mathbf{U} \},
\end{equation}
i.e., $X_{e}$ is a set of invariant states where feasible inputs exist for the system to remain in these states.
For the simple model of double integrator, this terminal constraint~\eqref{static constraint} is equivalent to~$v^i_K=\mathbf{0}_d$.

\begin{remark} \label{remark:terminal-constraint}
    Note that once the above constraint is enforced, the horizon length~$K$ can be extended to \emph{infinity} as the planned state $x^i_k$ is identical to $x^i_K$ if $u^i_{k-1} = u_e$ for $k >K$ where $u_e$ is the control input that satisfies $x^i_K= \textbf{A} x^i_K + \textbf{B} u_e$.
    Thus it is also called Infinite-horizon MPC (IMPC).
    Notably, the infinite horizon ensures that the planned trajectory is remained to be collision-free beyond horizon such that, as the horizon recedes, a feasible trajectory is already prepared.
	\hfill $\blacksquare$
\end{remark}

Furthermore, the objective function in~\eqref{eq:original-mpc-cost} considered in this work consists of two parts:
\begin{equation} \label{eq:C^i}
	C^i = C^i_w + C^i_p.
\end{equation}
The first part
\begin{equation} \label{eq:C^i_w}
	C^i_w = \sum_{j \neq i} \rho^{ij} (\frac{w^{i j} }{\epsilon}-\ln w^{i j})
\end{equation}
is related to the warning band in~\eqref{eq: MBVC definition}, $\rho^{ij} > 0$ is a designated parameter for deadlock resolution later.
Note that $\lim_{w^{ij} \rightarrow 0^+} C^i_w = + \infty$ and $\frac{\partial C^i_w}{\partial w^{ij}}|_{w^{ij}=\epsilon}=0$.
The second part
\begin{equation*}
	C^i_p = \frac{1}{2} Q_K \|p_{K}^{i}-p_{\text {target }}^{i}\|^2 + \frac{1}{2} \sum_{k=0}^{K-1} Q_k \|p_{k+1}^{i}-p_{k}^{i}\|^2 \\
\end{equation*}
is similar to the commonly-seen quadratic cost to penalize the distance to target and summed velocity,
where~$Q_k$, $k = 0, 1, \ldots, K$ are the weighting parameters.
In particular, $Q_k>0$, $k \in \mathcal{K}$ and  $Q_0 = 0$.
Given the above components, the optimization in~\eqref{eq:original-mpc} is rewritten as follows:
\begin{subequations} \label{eq:convex-program}
	\begin{align}
		&\min _{\boldsymbol{u}^{i}, \boldsymbol{x}^{i}, w^{i j} } C^{i}  \notag \\
		\textbf{ s.t. }\quad & {a_{k}^{i j}}^\mathrm{T} p_{k}^{i} \geq b_{k}^{i j}, \;\forall j\neq i,\;  k \in \tilde{\mathcal{K}}; \label{eq:convex-program-avoid-k}\\
		&{a_{K}^{i j}}^\mathrm{T} p_{K}^{i} \geq b_{K}^{i j} + w^{i j},\;\forall j\neq i;  \label{eq:convex-program-avoid-K} \\
		&\epsilon \geq w^{i j} \geq 0; \label{eq:convex-program-epsilon}\\
		&v^i_K=\mathbf{0}_d;  \label{eq:convex-program-terminal}\\
		&\eqref{eq:dynamic-constraint}-\eqref{eq:velocity-constraint}.\notag
	\end{align}
\end{subequations}
where~\eqref{eq:convex-program-avoid-k}-\eqref{eq:convex-program-epsilon} are the re-formulated constraints for collision avoidance and~\eqref{eq:convex-program-terminal} is the newly introduced terminal constraint;
and the objective function~$C^i$ is defined in~\eqref{eq:C^i}.

\subsection{Condition for Deadlocks} \label{subsection:deadlock-analysis}

The following theorem reveals that deadlock can only happen under specific conditions.

\begin{theorem} \label{theorem:deadlock-property}
 	If robot $i \in \mathcal{N}$ belongs to a deadlock, the following condition holds:
	\begin{equation} \label{eq:necessary-condition}
		Q_K \left(p_{\text{target}}^{i}-p^{i}_{K}\right)+ \sum_{j\in \mathcal{N}^i} \rho^{ij} \delta_{i j} a_{K}^{i j} =0,
	\end{equation}
	where~$ \mathcal{N}^i \triangleq \left\{ j \, | \, w^{i j} < \epsilon \right\}$, $\delta_{i j}=\frac{\epsilon-w^{i j}}{ \epsilon w^{i j} }$; 
	and $w^{ij} = w^{ji}$ holds for $j \in \mathcal{N}^i$.
\end{theorem}

\begin{proof}
	To begin with, constraint~\eqref{eq:dynamic-constraint} in~\eqref{eq:convex-program} can be directly expanded as $x_{k}^{i} = \mathbf{A}^{k} x_{0}^{i} + \mathbf{A}^{k-1}\mathbf{B}u_{0}^{i} + \cdots +\mathbf{B}u_{k-1}$, $k\in \mathcal{K}$.
	Moreover, the domain of $w^{ij}$ in $C^i$ is $w^{ij}>0$ since $\lim_{w^{ij} \rightarrow 0^+} C^i_w = +\infty$ and $C^i_w$ is only convex when $w^{ij} > 0$.
	Therefore, the constraints $w^{ij} \geq 0$ can be omitted in \eqref{eq:convex-program}.
	Then, the Lagrange function of~\eqref{eq:convex-program} is given as
	\begin{equation*}
		\begin{aligned}
			\mathcal{L}^{i}=&C^{i}+\sum_{k=1}^{K} {^{u}\lambda_{k}^{i}}\left(\|\Theta_a u_{k-1}^{i}\|-u_{\max }\right)\\
			&+\sum_{k=1}^{K} {^{v}\lambda_{k}^{i}}\left(\|\Theta_v v_{k}^{i}\|-v_{\max }\right) \\
			&+\sum_{j \ne i} \lambda_{K}^{i j}\left(b_{K}^{i j}+w^{i j}-{a_{K}^{i j}}^\mathrm{T} p_{K}^{i}\right) + \sum_{j \ne i} {}^w \lambda^{i j} (w^{i j}-\epsilon)\\
			&+\sum_{k=1}^{K-1} \sum_{j \ne i} \lambda_{k}^{i j}\left(b_{k}^{i j}-{a_{k}^{i j}}^\mathrm{T} p_{k}^{i}\right)
			+{^t \nu^{i}}^\mathrm{T} v_{K}^{i} \\
			&+\sum_{k=1}^{K} {\nu_{k}^{i}}^\mathrm{T} \left(x_{k}^{i}-\mathbf{A}^{k} x_{0}^{i}-\mathbf{A}^{k-1} \mathbf{B} u_{0}^{i}-\cdots- \textbf{B} u^i_{K-1}\right),
		\end{aligned}
	\end{equation*}
	where ${^{u}\lambda_{k}^{i}}$, ${^{v}\lambda_{k}^{i}}$, $\lambda_{k}^{i j}$, $\nu_{k}^{i}= \left[ {^{p}\nu_{k}^{i}},{^{v}\nu_{k}^{i}} \right] $ and ${}^t \nu^{i}$ are the Lagrangian multipliers to the corresponding inequality and equality constraints, respectively.

	When a deadlock happens, robot~$i$ remain static and thus~$u_{k-1}^{i}=\mathbf{0}_d$ and $v^{i}_{k}=\mathbf{0}_d$ hold.
	It implies that both $\|\Theta_a u_{k-1}^{i}\| < u_{ \text{max} } $ and $  \| v_{k}^{i}\| < v_{\text{max} }$ hold. 
	Hence, according to the complementary slackness condition of Karush-Kuhn-Tucker (KKT) condition \cite{Boyd2004}, ${^{u}\lambda_{k}^{i}} = 0$ and ${^{v}\lambda_{k}^{i}}=0$ hold.
	Furthermore, according to the stationary condition of KKT condition, the following equations are satisfied:
	\begin{subequations} \label{eq:KKT}
		\begin{align}
			&\frac{\partial \mathcal{L}^{i}}{\partial p_{k}^{i}}=\frac{\partial C^{i}}{\partial p_{k}^{i}}-\sum_{j \ne i } \lambda_{k}^{i j} a_{k}^{i j}+{ }^{p} \nu_{k}^{i}=0, \label{eq:KKT-p} \\
			&\frac{\partial \mathcal{L}^{i}}{\partial v_{k}^{i}}=\left\{\begin{array}{c}
				{}^v \nu_{k}^{i}, \quad k \in \tilde{\mathcal{K}} \\
				{}^v \nu_{k}^{i}+ {}^t \nu^{i}, \quad k=K
			\end{array}=0, \right. \label{eq:KKT-v} \\
			&\frac{\partial \mathcal{L}^{i}}{\partial u_{k-1}^{i}}=-\textbf{B}^{\mathrm{T}} {\mathbf{A}^{\mathrm{T}}}^{K-k} \nu_{K}^{i}
			-\textbf{B}^{\mathrm{T}} {\mathbf{A}^{\mathrm{T}}}^{K-k-1} \nu_{K-1}^{i}   \label{eq:KKT-u}\\
			& \qquad \qquad -\cdots-\textbf{B}^{\mathrm{T}} \nu_{k}^{i}=0. \notag \\
			&\frac{\partial \mathcal{L}^{i}}{\partial w^{i j}}=\frac{\partial C^{i}}{\partial w^{i j}}+\lambda_{K}^{i j} + {}^w \lambda^{i j}=0, \label{eq:KKT-w}
		\end{align}
	\end{subequations}
    Given the actual value of matrices~$\mathbf{A},\,\mathbf{B}$, condition~\eqref{eq:KKT-u} can be rewritten as
	\begin{equation} \label{eq:nu-AB}
		{\nu_{K}^{i}}^{\mathrm{T}} \mathbf{A}^{K-k} \textbf{B} + {\nu_{K-1}^{i}}^{\mathrm{T}} \mathbf{A}^{K-k-1} \textbf{B} + \cdots +{\nu_{k}^{i}}^{\mathrm{T}} \textbf{B} = 0,
	\end{equation}
	which directly implies that ${\nu_{K}^{i}}^{\mathrm{T}} \textbf{B} = 0$, i.e.,
	$  \left[ { {}^p \nu_{K}^{i}}^{\mathrm{T}},{{}^v \nu_{K}^{i}}^{\mathrm{T}} \right]^\mathrm{T}
	\left[
	\begin{array}{c}
		0 \\
		\mathbf{I}_d
	\end{array}
	\right]=0$
	and~${ }^{v} \nu_{K}^{i} =0$.
	Combined with condition~\eqref{eq:KKT-v}, ${}^{v} \nu_{k}^{i}=0$, $k \in \mathcal{K}$ holds. Then,~\eqref{eq:nu-AB} can be further simplified as
	\begin{equation*}
	(K-k) \ h \ {}^{p} {\nu_{K}^{i}} + (K-k-1) \ h\ {{}^{p} \nu_{K-1}^{i}}
			+ \cdots + h \ { }^{p} {\nu_{k+1}^{i}} =0.
	\end{equation*}
	By setting~$k=K-1$, it follows that~${{ }^{p} \nu_{K}^{i} }=0$.
    Consequently, \eqref{eq:KKT-p} implies that
	\begin{equation} \label{eq:LpK}
		\frac{\partial \mathcal{L}^{i}}{\partial p_{K}^{i}}=\frac{\partial C^{i}}{\partial p_{K}^{i}}-\sum_{ j \ne i } \lambda_{K}^{i j} a_{K}^{i j}=0.
	\end{equation}

	\begin{figure}[t!]
		\centering
		\includegraphics[width=0.8\linewidth]{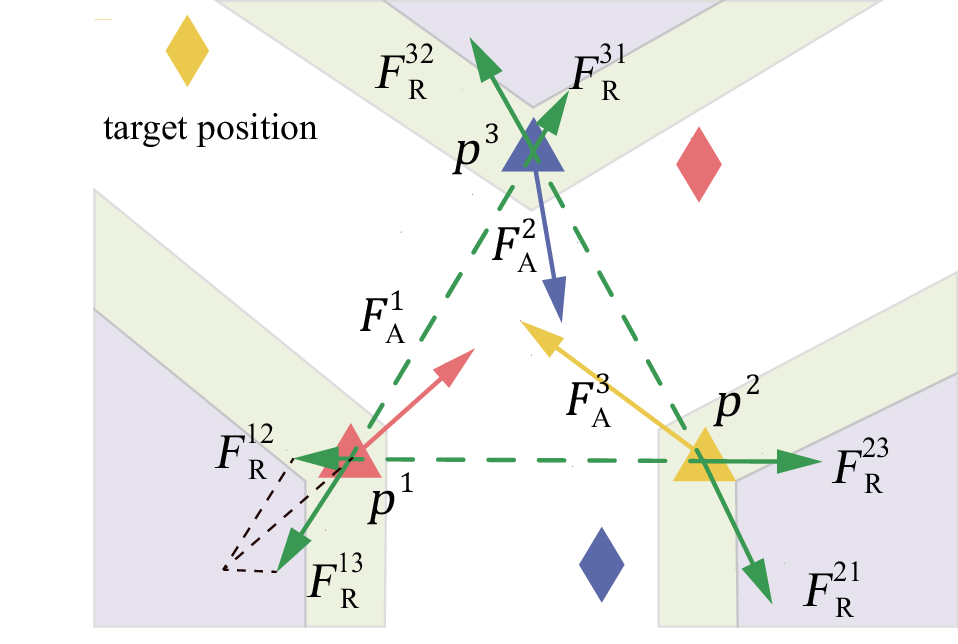}
		\caption{ Deadlock can be treated as a force equilibrium, where the attractive force from the target (in yellow, red, blue) and the repulsive forces from other robots (in green) are balanced. }
		\label{MBVC deadlock}
	\end{figure}

	Moreover, by substituting
	\eqref{eq:C^i} into \eqref{eq:KKT-w}, it follows that
	\begin{equation} \label{eq:KKT-w-2}
		\rho^{ij} (-\frac{1}{w^{i j}} + \frac{1}{\epsilon})+\lambda_{K}^{i j} + {}^w \lambda^{i j} =0.
	\end{equation}
	Assume that there exists $j \ne i$ such that ${}^w \lambda^{i j} > 0$ holds.
  	Due to the complementary slackness, $w^{i j}=\epsilon$ holds.
	Thus, \eqref{eq:KKT-w-2} implies~$\lambda_{K}^{i j} + {}^w \lambda^{i j} =0$.
	However, the KKT condition requires that
	both ${}^w \lambda^{i j} \ge 0$ and  $\lambda_{K}^{i j} \ge 0$ must hold.
	As a result, it follows that ${}^w \lambda^{i j} = 0$ and  $\lambda_{K}^{i j} = 0$ hold,
    which contradicts the assumption that ${}^w \lambda^{i j} > 0$.
	Thus, it holds that~${}^w \lambda^{i j} =0, \,   \forall j \ne i$.
	Furthermore, combined with~\eqref{eq:C^i_w} and \eqref{eq:KKT-w}, it follows that
	\begin{equation} \label{eq:lambda-ijk}
		\lambda_{K}^{i j}=-\frac{\partial C^{i}}{\partial w^{i j}}=-\rho^{ij} \frac{w^{i j}-\epsilon}{ \epsilon w^{i j} }.
	\end{equation}
	By substituting \eqref{eq:lambda-ijk} into \eqref{eq:LpK} and combining $p^i_K=p^i_{K-1}$, it can be derived that
	\begin{equation*}
		Q_K \left(p_{\text{target}}^{i}-p^{i}_{K}\right) + \sum_{j \ne i} a_{K}^{i j} \rho^{ij} \frac{\epsilon-w^{i j}}{ \epsilon w^{i j} }=0.
	\end{equation*}
	Moreover, if robot~$j\notin \mathcal{N}^i$, then both $w^{i j} = \epsilon$ and $\frac{w^{i j}-\epsilon}{ \epsilon w^{i j} }=0$ hold, which leads to equation~\eqref{eq:necessary-condition}.

	For $j \in \mathcal{N}^i$, since $w^{ij} < \epsilon$ and ${}^w \lambda^{i j} =0$, from equation~\eqref{eq:KKT-w-2}, it is clear that $\lambda^{i j}_k > 0$ which derives
	\begin{equation} \label{eq:aij-p-b-w}
		a_{K}^{i j^\mathrm{T}} p_{K}^{i} = b_{K}^{i j} + w^{i j}
	\end{equation}
	via complementary slackness condition.

	 {
		For robot $j \in \mathcal{N}^i$, it will be static as well as robot $i$; otherwise, due to the change of $\overline{p}^j_K$ and $a^{ij}_K$, the equalities~\eqref{eq:necessary-condition} and \eqref{eq:aij-p-b-w} cannot be constantly held.
		In the following, we will show that $w^{ji} < \epsilon$ by contradiction.
		Considering that $w^{ji} \leq \epsilon$, assume $w^{ji} = \epsilon$ such that ${a_K^{ji}}^\mathrm{T} p_K^j \geq b_K^{ji} + \epsilon$  holds.
		Combining this with equation~\eqref{eq:aij-p-b-w} yields 
		\begin{equation} \label{eq:w^{ij}-epsilon}
			b_K^{ij} +w^{ij} - {a_K^{ij}}^\mathrm{T} p_K^i + {a_K^{ji}}^\mathrm{T} p_K^j - b_K^{ji} -\epsilon \geq 0.
		\end{equation} 
		When robots $i$ and $j$ remain static, as both $\overline{p}_K^i = p_K^i(t-h)=p_K^i$ and $\overline{p}_K^j = p_K^j(t-h)=p_K^j$ hold, it can be derived that $a_K^{ij} = \frac{ p_K^{ij} }{ \| p_K^{ij} \| }$ and $b_K^{ij} = {a^{i j}_K}^\mathrm{T} \frac{p_{K}^{i} + p_{K}^{j}}{2}+\frac{r_{\min }^{\prime}}{2}$ also hold.
		Then, $a_K^{ji} = \frac{ p_K^{ji} }{ \| p_K^{ji} \| }$ and $b_K^{ji} = {a^{ji}_K}^\mathrm{T} \frac{p_{K}^{j} + p_{K}^{i}}{2}+\frac{r_{\min }^{\prime}}{2}$ can be similarly derived.
		Substituting them into inequality~\eqref{eq:w^{ij}-epsilon}, we can obtain that $w^{ij} -\epsilon \geq 0$, which contradicts with the fact that $w^{ij} < \epsilon$.
		Therefore, $w^{ji} <\epsilon$ must hold.
	}

	Moreover, as robot~$j$ remains static, similar to aforementioned steps, ${}^w \lambda^{ji} = 0$ can be derived.
	Combining it with $w^{ji} <\epsilon$, we can obtain that 
	\begin{equation} \label{eq:aji-p-b-w}
		{a_{K}^{ji}}^\mathrm{T} p_{K}^j = b_{K}^{ji} + w^{ji}.
	\end{equation}
	By substituting $a^{i j}_K$ and $b^{i j}_K$ in addition to combining \eqref{eq:aij-p-b-w} and \eqref{eq:aji-p-b-w}, we can get that $w^{ij}=w^{ji}$.
	This completes the proof.
\end{proof}

Based on Theorem~\ref{theorem:deadlock-property}, $F^i_A=Q_K \left(p_{\text{target}}^{i}-p^{i}_{K}\right)$ can be regarded as the attractive force from the target and $ F_R^{ij}=  \rho^{ij}  \delta_{i j} a^{i j}_{K}$  is the repulsive force from robot~$j$, where $a^{i j}_{K}$ and $\rho^{ij}  \delta_{i j} $ are the direction and magnitude, respectively.
As illustrated in Fig.~\ref{MBVC deadlock}, the deadlock condition~\eqref{eq:necessary-condition} can be understood as a balance of these forces, i.e., $F^i_{A} +\sum_{j \ne i}  F_{R}^{ij} =0$.

\begin{remark} \label{remark: deadlock contrast}
  	Without the warning band, i.e., $w^{i j}=0$ and when the objective function~$C^i$ only includes~$C^i_p$ in~\eqref{eq:C^i}, the necessary condition of deadlocks is rewritten as
	\begin{equation}\label{eq:simple-deadlock-cond}
		Q_K \left(p_{\text {target}}^{i}-p_{K}^{i}\right) + \sum_{j \ne i} \lambda_{K}^{i j} a_{K}^{i j}=0,
	\end{equation}
	which indicates that the magnitude of the repulsive force~$\lambda_{K}^{i j}$ will be passively determined.
	In comparison, by introducing the warning band variable~$w^{ij}$ and including it in the objective function~\eqref{eq:C^i_w}, the magnitude of the repulsive force satisfies \eqref{eq:lambda-ijk} and contains $\rho^{ij}$ as a parameter.
	Thus, for deadlock resolution, the repulsive forces can be actively adjusted by changing $\rho^{ij}$,
    rather than being passively determined as in~\eqref{eq:simple-deadlock-cond}.
    In other words, via adjusting~$\rho^{ij}$ properly, the condition in~\eqref{eq:necessary-condition} can be falsified to resolve potential deadlocks.
	\hfill $\blacksquare$
\end{remark}

\begin{figure}[t]
	\centering
	\includegraphics[width=0.6\linewidth]{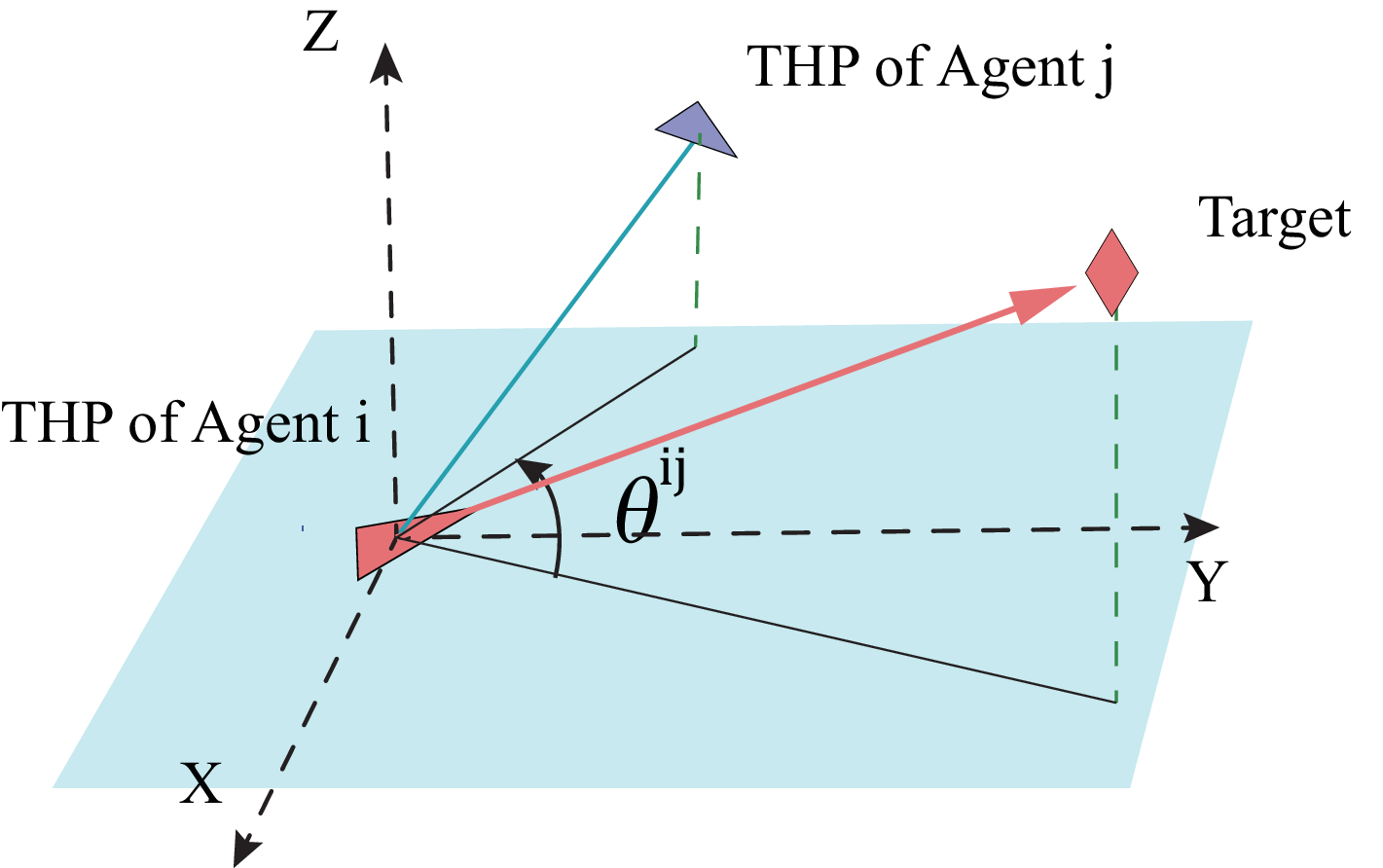}
	\caption{Illustration of how~$\theta^{i j}$ in~\eqref{eq: rho^{ij}} is computed,
          given the terminal horizon position (THP) of robots~$i,j$ and the target.}
	\label{figure:theta}
\end{figure}

\begin{remark}
It should be mentioned that similar analyses based on KKT conditions can be found in \cite{Grover2020,Grover2020-1,Grover2022}, which utilizes a control barrier function method.
By contrast, the analysis in this theorem is built upon an MPC with predicted horizon, and many additional constraints such as \eqref{eq:convex-program-avoid-k}-\eqref{eq:convex-program-terminal} and a new term \eqref{eq:C^i_w} related to the warning band is introduced in the optimization.
Furthermore, different from \cite{Grover2020-1,Grover2022}, the warning band above provides elastic safety margin when possible deadlock happens.
\hfill $\blacksquare$
\end{remark}

\subsection{ Deadlock Detection and Resolution} \label{subsection:deadlock resolution}

A potential deadlock is detected when the following condition termed \emph{terminal overlap} holds.

\begin{definition} [Terminal Overlap] \label{def:terminal-overlap}
	The terminal overlap for robot~$i$ happens when~$p^i_K(t)=p^i_K(t-h)$, $p^i_K(t) \ne p^i_\text{target}$, $p^i_K(t)=p^i_{K-1}(t)$, and $p^i_{K-1}(t)=p^i_{K-2}(t)$ hold.
	\hfill $\blacksquare$
\end{definition}

The above condition can be analyzed as follows.
According to the condition of deadlocks in~\eqref{eq:necessary-condition}, the summed forces at the planned terminal position~$p^i_K$ equals to zero and remains so indefinitely.
Consequently, the planned terminal position at two consecutive time steps are identical,
i.e., $p^i_K(t)={p^i_K(t-h)}$ holds.
Furthermore, the planned positions at the preceding steps of horizon would approach the same position as time evolves and eventually~$p^i_k$ overlaps for~$k=K, K-1, K-2$.
In other words, the condition of terminal overlap above allows an early detection of potential deadlocks by time~$(K-2)h$.
A duration longer than~$2$ steps can also be adopted.
However, longer duration indicates later detection of deadlocks.

After a positive detection, a resolution scheme is proposed by adapting the coefficient~$\rho^{ij}$ as follows:
\begin{equation} \label{eq: rho^{ij}}
  \begin{aligned}
	\rho^{ij} &= \rho_0\, e^{\left( \eta^i(t) \, \sin \theta^{i j}\right)},\\
	\eta^i(t) &= \left\{
				\begin{array} {rrll}
		            &\eta^i(t-h) + \Delta \eta,  &\mbox{if}\; b^i_\text{TO} = True; \\
		            &0,                          &\mbox{if}\; \ w^{i j} = \epsilon, \forall j \ne i;\\
		            &\eta^i(t-h),                &\text{otherwise};
				\end{array}	\right.
\end{aligned}
\end{equation}
where $\rho_{0} > 0$ and $\Delta \eta >0$ are parameters; initially~$\eta^i(t_{0})=0$; $b^i_\text{TO} = True$ holds when a terminal overlap happens; $\theta^{i j}$  is defined as the angle in $xy$ plane between the projection of line $\overline{p}^i_K$  to its target position $p^i_{\text{target}}$ and to $\overline{p}^j_K$, as illustrated in Fig.~\ref{figure:theta}.

\begin{figure}[t]
	\centering
	\includegraphics[width=1.0\linewidth]{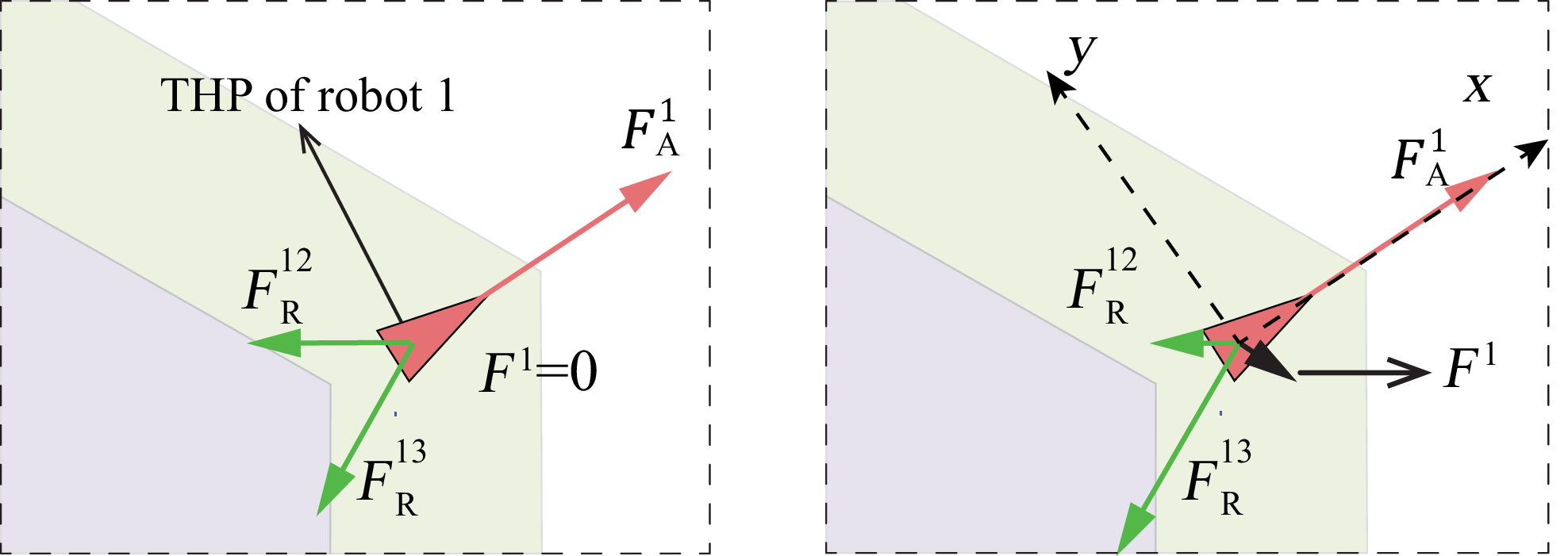}
	\caption{
		\textbf{Left}: The condition of deadlocks is a force equilibrium for robot~$1$ in which the resulting force $F^1=0$.  
		\textbf{Right}: After introducing the right-hand rule, the repulsive forces to the left side $F^{13}_R$ is increased while the force to the right side $F^{12}_R$ is decreased gradually, yielding the summed force $F^{1}$ nonzero. 
		Thus the condition of deadlocks is falsified.
	}
	\label{figure:deadlock-resolve}
\end{figure}

Consequently, $\rho^{ij}$ is adapted to modify the repulsive forces by following the proposed right-hand rule.
Initially, $\rho^{ij} = \rho_0$ and no right-hand forces are introduced.
If $b^i_\text{TO} = True$, $\eta^i$ is increased and becomes positive.
Consider the following two cases:
when robot~$j$ is on the left-hand side of robot~$i$, i.e.,  $\theta^{i j} > 0$  and thus $e^{\left( \eta^i(t) \, \sin \theta^{i j}\right)} > 1$.
As a result, the repulsive force from robot~$j$ is increased and thus robot~$i$ moves further away from robot~$j$;
Secondly, when robot~$j$ is on the right-hand side, robot~$i$ approaches to it.
Once $w^{i j} = \epsilon$, $\forall j \ne i$ holds, it is evident that $p^i_K$ must be outside the warning band of any others and there exists no $j \in \mathcal{N}^i$.
Then, $\eta^{i}$ returns to the initial value~$0$.
Otherwise, $\eta^{i}$ remains unchanged.

\begin{remark}
	Compared with other methods proposed in~\cite{Wang2017,Zhou2017},
	deadlocks can only be detected \emph{after} they happen and thus subsequently resolved ``on the spot''.
	Via the detection mechanism described above,
	deadlocks can be detected and immediately resolved earlier, \emph{before} they actually happen.
	As also validated by the numerical results in the sequel,
	this can lead to a safer and much more efficient navigation scheme, especially in crowded and high-speed scenarios.
	\hfill $\blacksquare$
\end{remark}

\subsection{The Complete Algorithm} \label{IMPC-DR algorithm}

The complete method is summarized in Algorithm~\ref{AL:IMPC-DR}.
When the system starts, the predetermined trajectory is initialized in Line~\ref{algline:intial} as $\overline{\mathcal{P}}^i(t_0)=\left [ p^{i}(t_0), \ldots, p^{i}(t_0)\right ]$.
The main loop in Algorithm~\ref{AL:IMPC-DR} runs as follows.
First, the predetermined trajectory~$\overline{\mathcal{P}}^i$ of each robot as defined in Definition~\ref{def:pt} is communicated with its neighboring robots in Line~\ref{algline:communication}.
Afterwards, the constraints for collision avoidance $cons^i$ in~\eqref{eq: a p > b} are derived via MBVC-WB in Line~\ref{algline:get-constraint}.
Thereafter, the important coefficient $\rho^{ij}$ adopted to the deadlock resolution is obtained.
The optimization~\eqref{eq:convex-program} is formulated with the updated constraints and solved subsequently in Line~\ref{algline:convex-programming}.
Based on the result, the boolean variable related to deadlock detection~$b^i_\text{TO}$ is obtained in Line~\ref{algline:get-boolean}.
Afterwards, the predetermined trajectory is derived from the planned one via its definition as well.
Finally, the planned trajectory is executed by the low-level tracking controller of each robot.
The above procedure repeats itself until all robots have reached their target positions.

The computational complexity of the proposed algorithm is analyzed as follows.
Since robot~$i$ communicates with other robots in~$\mathcal{N}$, its corresponds to~$(K+1)\cdot N + K\cdot d$ constraints in~\eqref{eq:convex-program} and~$(K\cdot d + N)$ real variables.
The convex programming~\eqref{eq:convex-program} at each time step can be solved in polynomial time w.r.t. the problem size, e.g., using the interior-point method.
Moreover, as shown in the sequel, each robot only needs to communicate with others robot within a certain range, this further reduces the computation complexity above.
The complexity of other components, such as the derivation of constraints for collision avoidance and deadlock resolution, are neglected compared with solving the optimization above.

\begin{algorithm}[t] \label{AL:IMPC-DR}
	\caption{IMPC-DR}
	\SetKwInOut{Input}{Input}\SetKwInOut{Output}{Output}

	\Input{$p^i(t_0)$, $p^i_\text{target}$}
	$b^i_\text{TO} \leftarrow  \textbf{False}$\;
	$\overline{\mathcal{P}}^i(t_0) \leftarrow \text{InitialPT}(p^i(t_0))$  \label{algline:intial}\;
	\While{not all robots at target}
	{
		\For{ $i\in \mathcal{N}$ concurrently }{
		  $\overline{\mathcal{P}}^j (t) \leftarrow \text{Communicate}(\overline{\mathcal{P}}^i(t))$\ \label{algline:communication}\;
                  $cons^i \leftarrow \text{MBVC-WB}(\overline{\mathcal{P}}^i(t), \overline{\mathcal{P}}^j (t))$ \label{algline:get-constraint}\;
                  	$\rho^{ij} \leftarrow \text{DeadlockResolve}()$\;
		  $\mathcal{P}^i(t) \leftarrow \text{Optimization}(cons^i,\rho^{ij})$ \label{algline:convex-programming}\;
                  $b^i_\text{TO} \leftarrow \text{DeadlockDetction}( \mathcal{P}^i(t))$ \label{algline:get-boolean}\;
			$\overline{\mathcal{P}}^i(t+h) \leftarrow \text{GetPT}( \mathcal{P}^i(t) )$\;
			$\text{ExecuteTrajectory}(\mathcal{P}^i(t))$ \label{algline:execute-trajectory}\;
		}
		$t \leftarrow t+h$\;
	}
\end{algorithm}

\section{Property Analyses} \label{section:analysis}

This section is devoted to the property analyses of the proposed algorithm,
regarding deadlocks, feasibility, and local communication.

 {
	Before moving forward, we first define the stability and instability of deadlocks in the 
	sense of Lyapunov  \cite{Slotine1991}.
}

\begin{definition} [Stable Deadlock] \label{def:stable-deadlcok}
	 {
		The deadlock equilibrium $x^i_{d}$ of robot~$i$ is stable, if for any $R>0$, there exists $r>0$ such that if $\|x^i(t_d)-x^i_{d} \| <r$, then $\|x^i(t)-x^i_{d} \| <R$ for any $t \geq t_d$, where $t_d$ is defined as in Definition 2. Otherwise, the deadlock equilibrium is unstable.
	}
\end{definition}
 {
	Note that for the cases where the multi-robot system has multiple deadlock equilibria, any deadlock equilibrium cannot be globally stable or unstable.
}

\begin{theorem} \label{theorem:deadlock-free}
	Assuming that
	(i) $\| p^i_\text{target} - p^j_\text{target} \| > r^\prime_\text{min}+2\epsilon$
	holds for $\forall i \ne j$;
	and (ii) projections of the targets of three or more robots onto the horizontal plane are not collinear, holds, no stable deadlocks exist under Algorithm~\ref{AL:IMPC-DR}.
\end{theorem}

\begin{figure}[t]
	\centering
	\includegraphics[width=1.0\linewidth]{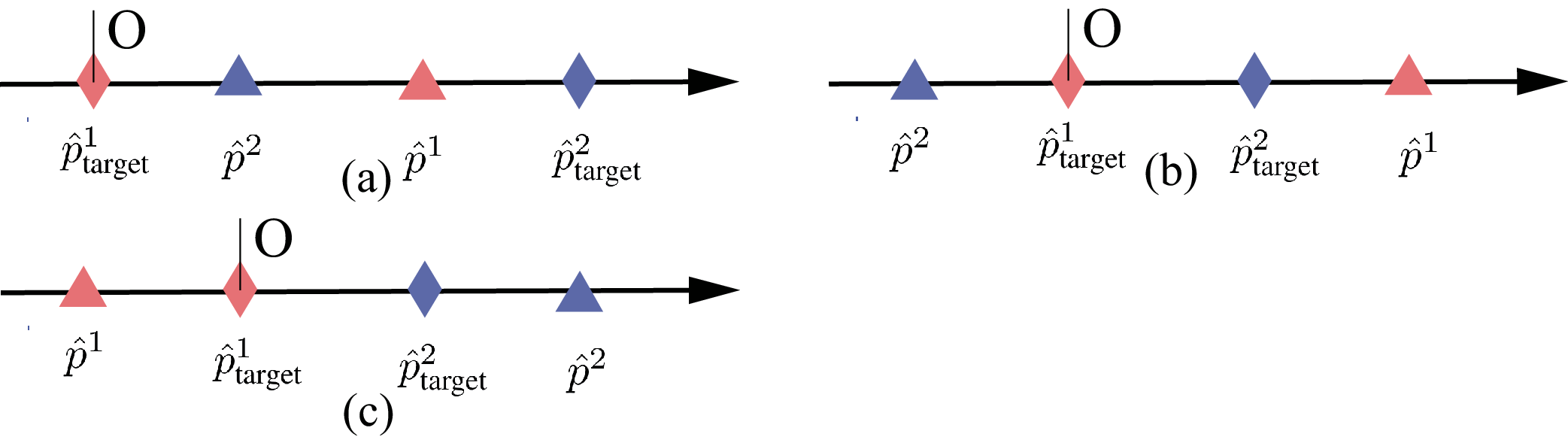}
	\caption{
		Illustration of three possible cases when two robots form a deadlock.
		}
	\label{figure:2_deadlock}
\end{figure}

\begin{proof}
	Once the condition of terminal overlap in Definition~\ref{def:terminal-overlap} holds,
	$\rho^{ij} = \rho_0\, e^{\left( \eta^i(t-h)\, \sin \theta^{i j}\right)}$ is replaced by
	$\rho^{ij} = \rho_0\, e^{\left( \eta^i(t) \sin \theta^{i j}\right)}$ where $\eta^i(t) = \eta^i(t-h) + \Delta \eta$.
	Now consider the direction from robot~$i$ to its target position as the $x$-axis and its orthogonal line as the $y$-axis, as illustrated in Fig.~\ref{figure:deadlock-resolve}.
	Assume that this deadlock holds indefinitely.
	By Theorem~\ref{theorem:deadlock-property}, the summed forces acting on robot~$i$ in~$y$-direction are given by
	\begin{equation} \label{F^i_{x y}}
		F^i_y= \sum_{j\in \mathcal{N}^i} (-\sin \theta^{i j})  \rho_{0} e^{\left( \eta^i(t) \sin \theta^{i j}\right)} \delta^{i j} = 0.
	\end{equation}
	Meanwhile, since this deadlock persists and $\theta^{ij}$, $\delta^{ij}$ do not change, the equilibrium condition similar to \eqref{F^i_{x y}} already holds at time $t-h$, i.e.,
	\begin{equation} \label{F^i_{A,xy}}
		\sum_{j\in \mathcal{N}^i} (-\sin \theta^{i j})   \rho_{0}  e^{\left( \eta^i(t-h) \sin \theta^{i j}\right)} \delta^{i j} = 0.
	\end{equation}
	Combining~\eqref{F^i_{x y}} and \eqref{F^i_{A,xy}}, it follows that
	\begin{equation*}
		F^i_{y} = \sum_{j\in \mathcal{N}^i} \sin \theta^{i j} ( 1- e^{\left( \Delta \eta \sin \theta^{i j}\right)} ) \rho_{0} e^{\left( \eta^i(t-h) \sin \theta^{i j}\right)} \delta^{i j}.
	\end{equation*}
	Since~$\sin \theta^{i j} ( 1-  e^{\left( \Delta \eta \sin \theta^{i j}\right)} ) \leq 0$
	holds for any~$\theta^{ij}\in (-\pi,\, \pi]$,
	it follows that $F^i_{y} \leq 0$.
	Moreover, the equality $F^i_{y} = 0$ holds if and only if $\theta^{ij} = 0$ or $\theta^{ij} = \pi$ holds, $\forall j \in \mathcal{N}^i$.

	Denote by $\mathbb{D}^i = \left\{ j \ | \ j=i \ {\rm or} \ j \in \mathcal{N}^i \ {\rm or} \ \exists k \in \mathcal{N}^j, k \in \mathbb{D}^i \right\}$ the set of robots within the same deadlock of robot $i$.
	For robot $j \in \mathbb{D}^i$, it also holds that $\theta^{jk} = 0$ or $\theta^{jk} = \pi$,
	$\forall k \in \mathcal{N}^j$.
	In other words, the projections of $p^j_K$ and $p^j_\text{target}$ onto the $xy$-plane are collinear, $\forall j \in \mathbb{D}^i$.
    Since the second condition in Theorem~\ref{theorem:deadlock-free} states that
	the projections of three or more robots' target are not collinear, the analyses can be focused on the case of two robots.
	Without loss of generality, these two robots are indexed by $1$ and $2$.
	By Lemma~\ref{lemma:deadlock} in the Appendix~\ref{appendix:deadlock}, their positions and targets are collinear.
	Therefore, as depicted in Fig.~\ref{figure:2_deadlock}, a new coordinate can be constructed with the line from $p^1_\text{target}$ to $p^2_\text{target}$ being the $x$-axis and $p^1_\text{target}$ being the origin.
	In this new coordinate, their positions are denoted by $\hat{p}^1$ and $\hat{p}^2$, and their targets by~$\hat{p}^1_\text{target}$ and $\hat{p}^2_\text{target}$, respectively.

	Following Theorem~\ref{theorem:deadlock-property},
	$Q_K \left(p_{\text{target}}^{1}-p^{1}_{K}\right)+ \rho^{12} \delta_{12} a_{K}^{12} =0$
	holds for robot~$1$; and
	$Q_K \left(p_{\text{target}}^{2}-p^{2}_{K}\right)+ \rho^{21} \delta_{21} a_{K}^{21} =0$
    holds for robot~$2$.
	Additionally, $w^{12}=w^{21}$ and $\delta_{12} = \delta_{21}$ hold.
	Moreover, since both $\theta^{12}$ and $\theta^{21}$ are either $0$ or $\pi$,
    it holds that $\rho^{12}=\rho_0=\rho^{21}$.
	Then, since $a_{K}^{21}=-a_{K}^{12}$ holds, 
	$\left(p_{\text{target}}^{1}-p^{1}_{K}\right)+\left(p_{\text{target}}^{2}-p^{2}_{K}\right)=0$ holds,
    which leads to the following three cases as shown in Fig.~\ref{figure:2_deadlock}.

	Case (a): As proven in Lemma~\ref{lemma:deadlock}, the underlying deadlock is unstable for robot~1 and 2.

	Case (b): Since
	$\|p_{K}^{1}-p_{K}^{2}\| = \hat{p}^1 - \hat{p}^2 \geq \hat{p}^2_\text{target} - \hat{p}^1_\text{target} > r^\prime_\text{min} + 2 \epsilon$ holds and $w^{12}, \ w^{21} \leq \epsilon$,
    it can be verified that ${a_{K}^{12}}^\mathrm{T} p_{K}^{1} > b_{K}^{12} + w^{12}$ and ${a_{K}^{21}}^\mathrm{T} p_{K}^{2} > b_{K}^{21} + w^{21}$ are satisfied.
	According to complementary slackness condition, this yields that $\lambda^{12}_K=\lambda^{21}_K=0$ holds.
    Via~\eqref{eq:lambda-ijk}, this further indicates the repulsive force between robots $1$ and $2$ are zero, i.e., the robots are attracted by their targets only.

	Case (c): Clearly,
	$\hat{p}^1 \leq \hat{p}^1_\text{target}$
	and
	$\hat{p}^2 \geq \hat{p}^2_\text{target}$ hold.
    Thus, similar to the previous case,
    it implies that~$\hat{p}^2 - \hat{p}^1 > r^\prime_\text{min} + 2 \epsilon$
    and the condition for deadlocks cannot hold.

	To summarize, no stable deadlocks exists under the given two conditions.
\end{proof}

\begin{remark}
    These two conditions in Theorem \ref{theorem:deadlock-free} are not restrictive as the first condition requires that the targets are at least separated by the minimum safety distance; and the second condition that more than three targets are strictly collinear is quite rare to happen and can be easily avoided by slightly adjusting the target positions. 
    Moreover, as shown in Theorem~\ref{theorem:deadlock-free}, the only possible deadlocks are unstable, meaning that any slight deviations would
drive the system away from the deadlock equilibrim.
\hfill $\blacksquare$
\end{remark}

\begin{remark}
	The above theorem provides a theoretical guarantee for the proposed deadlock resolution scheme.
	In comparison, the artificial right-hand perturbation introduced in~\cite{Wang2017} and the right-hand detour points proposed in~\cite{Pierson2020, Abdullhak2021} are heuristic and thus lacking theoretical analyses.
	In addition, the above methods in general introduce an instantaneous change of the control inputs whenever deadlocks happen,
	while the proposed scheme adapts a smooth and gradual adaptation of repulsive forces, before the potential deadlocks actually happen.
	Moreover, a decentralized control strategy is proposed in \cite{Grover2020-1,Grover2022},
	which can provably drive the robots out of any possible deadlock.
    It unfortunately can only be applied to less than three robots and thereby lacks the generality of the proposed deadlock resolution in this paper.
\hfill $\blacksquare$
\end{remark}

\begin{theorem}\label{recursive feasible}
	The optimization in~\eqref{eq:convex-program} is recursively feasible under Algorithm~\ref{AL:IMPC-DR}.
\end{theorem}

\begin{proof} \label{proof of recursive feasibility}
	Given a feasible solution $u^i_{k-1}(t-h)$ and $x^i_k(t-~h)$ at time~$t-h$, $k\in \mathcal{K}$,
	it is proven in the sequel that~$x^i_k(t)=x^i_{k+1}(t-h)$, $u^i_{k-1}(t)=u^i_k (t-h)$, $k \in \mathcal{K}$ and 
	$w^{i j} (t) = \min \{ \epsilon, \; {a_{K}^{i j}}^\mathrm{T}(t) p_{K}^{i}(t-h) - b_{K}^{i j}(t) \}$ is a feasible solution at time~$t$. As mentioned in Remark~\ref{remark:terminal-constraint}, it is enforced that~${x^i_{K+1}(t-h)}=x^i_{K}(t-h)$ and $u^i_K(t-h)=u_e$.

	First, as the result of optimization at time step $t-h$, $x^i_{k+1}(t-h)$ and $u^i_{k}(t-h)$ with $k\in \tilde{\mathcal{K}}$, satisfy the constraints in~\eqref{eq:dynamic-constraint}-\eqref{eq:velocity-constraint} naturally.
	In addition, since $x^i_{K}(t)={x^i_{K+1}(t-h)}={x^i_{K}(t-h)}$ and $u^i_{K-1}(t)=u^i_K(t-h)=u_e$ hold, $x^i_{K}(t)$ and $u^i_{K-1}(t)$ also satisfy these constraints.
	In the meantime, as $x^i_{K}(t)=x^i_{K+1}(t-h)=x^i_{K}(t-h)=x^i_{K-1}(t)$ holds, it is evident that the constraint \eqref{eq:convex-program-terminal} holds as well.
	Second, it has been shown that, by substituting \eqref{a-b-definition}, \eqref{eq:convex-program-avoid-k} is equal to
	\begin{equation} \label{pj pj > r min 1}
		\| p_{k+1}^{i}(t-h) -p_{k+1}^{j}(t-h) \| \geq r^\prime_{\min }.
	\end{equation}
	As a feasible solution at time~$t-h$,  $p_{k+1}^{i}(t-h)$ satisfies
	${a_{k+1}^{i j}}^\mathrm{T}(t-h)\, p_{k+1}^{i}(t-h) \geq b_{k+1}^{i j}(t-h), ~\forall j \neq i, k\in \tilde{\mathcal{K}}$.
	Combined with Lemma~\ref{lemma:MBVC}, the inequality in~\eqref{pj pj > r min 1} holds as well as constraint \eqref{eq:convex-program-avoid-k}. 
	Lastly, it remains to be shown that the constraints in~\eqref{eq:convex-program-avoid-K}, ~\eqref{eq:convex-program-epsilon} and $w^{i j} > 0$ hold.
	As a feasible solution at time~$t-h$, $p_{K}^{i}(t-h)$ satisfies
	${a_{K}^{i j}}^\mathrm{T}(t-h) p_{K}^{i}(t-~h) \geq b_{K}^{i j}(t-h) + w^{i j}$,
	for $\forall j \neq i$.
	Since $w^{i j}(t-h)>0$ holds, it implies that ${a_{K}^{i j}}^\mathrm{T}(t-h) p_{K}^{i}(t-h) > b_{K}^{i j}(t-h)$ and further
	\begin{equation} \label{eq: pi-pj > r_min}
		\| p_{K}^{i}(t-h) -p_{K}^{j}(t-h) \| > r^\prime_{\min }.
	\end{equation}
	In addition, since~$p_{K}^{i}(t-h)=p_{K+1}^{i}(t-h)=p_{K}^{i}(t)$ holds, it yields
	$\| p_{K}^{ij}(t)\| > r^\prime_{\min }.$
	In combination with ${a_{K}^{i j}}(t)$ and $ b_{K}^{i j}(t)$ from~\eqref{a-b-definition},
    it is clear that
	\begin{equation} \label{ap>b}
		{a_{K}^{i j}}^\mathrm{T}(t) p_{K}^{i}(t) > b_{K}^{i j}(t).
	\end{equation}
	Since $w^{i j} (t) =\min \{ \epsilon, \; {a_{K}^{i j}}^\mathrm{T}(t) p_{K}^{i}(t) - b_{K}^{i j}(t) \}$,
    it implies that $w^{i j}(t)>0$.
    Thus, the proposed solution~$x^i_k(t)=x^i_{k+1}(t-h)$, $u^i_{k-1}(t)=u^i_{k}(t-h)$, $k\in \mathcal{K}$ and $w^{i j} (t) = \min \{ \epsilon, \; {a_{K}^{i j}}^\mathrm{T}(t) p_{K}^{i}(t) - b_{K}^{i j}(t) \}$ is a feasible solution at time~$t$.

    Consequently, if the initial optimizations are feasible for all agents, then they will be feasible in a recursive way and the recursive feasibility of optimization~\eqref{eq:convex-program} is ensured for all robots.
\end{proof}

\begin{remark}
It should be mentioned that the infeasible problem in motion planning was considered in previous related work, e.g., \cite{Park2022}.
Different from~\cite{Park2022} where the planned trajectory is represented by a Bezier-spline,
 this work considers a more general representation as a sequence of sampled points.
 Thus, the feasibility results above can be extended to more complex dynamics such as unicycles. More importantly, Theorem 2 together with Theorem 3 demonstrate that
 the proposed Algorithm~\ref{AL:IMPC-DR} indeed solves the deadlock resolution issue with feasibility guarantee.
 \hfill $\blacksquare$
\end{remark}

\begin{corollary}
  If  the optimization~\eqref{eq:convex-program} is initially feasible,
  the proposed algorithm ensures that all robots remain collision-free at all time.
\end{corollary}

\begin{proof}
  Since optimization~\eqref{eq:convex-program} is evidently feasible at~$t=t_0$,
  Theorem~\ref{recursive feasible} shows that it remains feasible for all future time.
  As an important constraint~\eqref{eq:convex-program-avoid-k}, the inter-robot collision avoidance
  is satisfied by any feasible solution.
  In other words, inter-robot collisions are avoided for all robots at all time.
\end{proof}

\begin{theorem} \label{theorem:communication-distance}
	 {
		Even if robot~$i\in \mathcal{N}$ only communicates with robot~$j\in \mathcal{N}$ that satisfies
		$\| p^{ij}\| \leq 2 v_{\text{max}} K h+r^{\prime}_{\text{min}}+2 \epsilon$, the properties mentioned in Theorems~\ref{theorem:deadlock-free} and ~\ref{recursive feasible} still hold.
	}
\end{theorem}

\begin{proof}
	To begin with, since $ \| p^{ij}\| \ge  2v_{\text{max}} K h+r^{\prime}_{\text{min}}+2 \epsilon$ holds,
	$ \| p^{ij}_k\| \ge  r^{\prime}_{\text{min}}+2 \epsilon$ holds, $\forall k\in \mathcal{K}$.
	Thus, there are no repulsive forces between robots~$i,\,j$,
	i.e., no deadlocks can appear.
	Second, constraints might be added or removed in~\eqref{eq:convex-program}
	if robots enter or leave the above range, respectively.
	More specifically,
	when an existing constraint is removed,
	its feasibility remains unchanged as the problem is less constrained.
	On the other hand, a new constraint is added
	when robot~$j$ enters the communication range of robot~$i$.
	Since $ \| p^{ij}\| \ge  2v_{\text{max}} K h+r^{\prime}_{\text{min}} + 2\epsilon$ holds,
	it follows that~$ \| p^{ij}_k\| \ge  r^{\prime}_{\text{min}}$ holds, $k\in \mathcal{K}$,
	and in turn~\eqref{pj pj > r min 1} holds.
	Following a similar proof of Theorem~\ref{recursive feasible},
	it can be shown that~$x^i_k(t)=x^i_{k+1}(t-h)$ and $u^i_{k-1}(t)=u^i_{k}(t-h)$
	is a feasible solution as well.
	Consequently, enforcing the local communication range above does not effect the
	theoretical guarantee of the complete algorithm on the deadlock-free property and recursive feasibility.
\end{proof}

\begin{remark}
  Many related work~\cite{Park2022, Jesus2021} requires a fully-connected communication network.
  By contrast, the above theorem shows that the proposed algorithm requires a local communication strategy, i.e., each robot only communicates with other robots that lie within a communication range.
  \hfill $\blacksquare$
\end{remark}

\begin{figure}[t]
	\centering
	\subfigure{
		\includegraphics[width=1.0\linewidth]{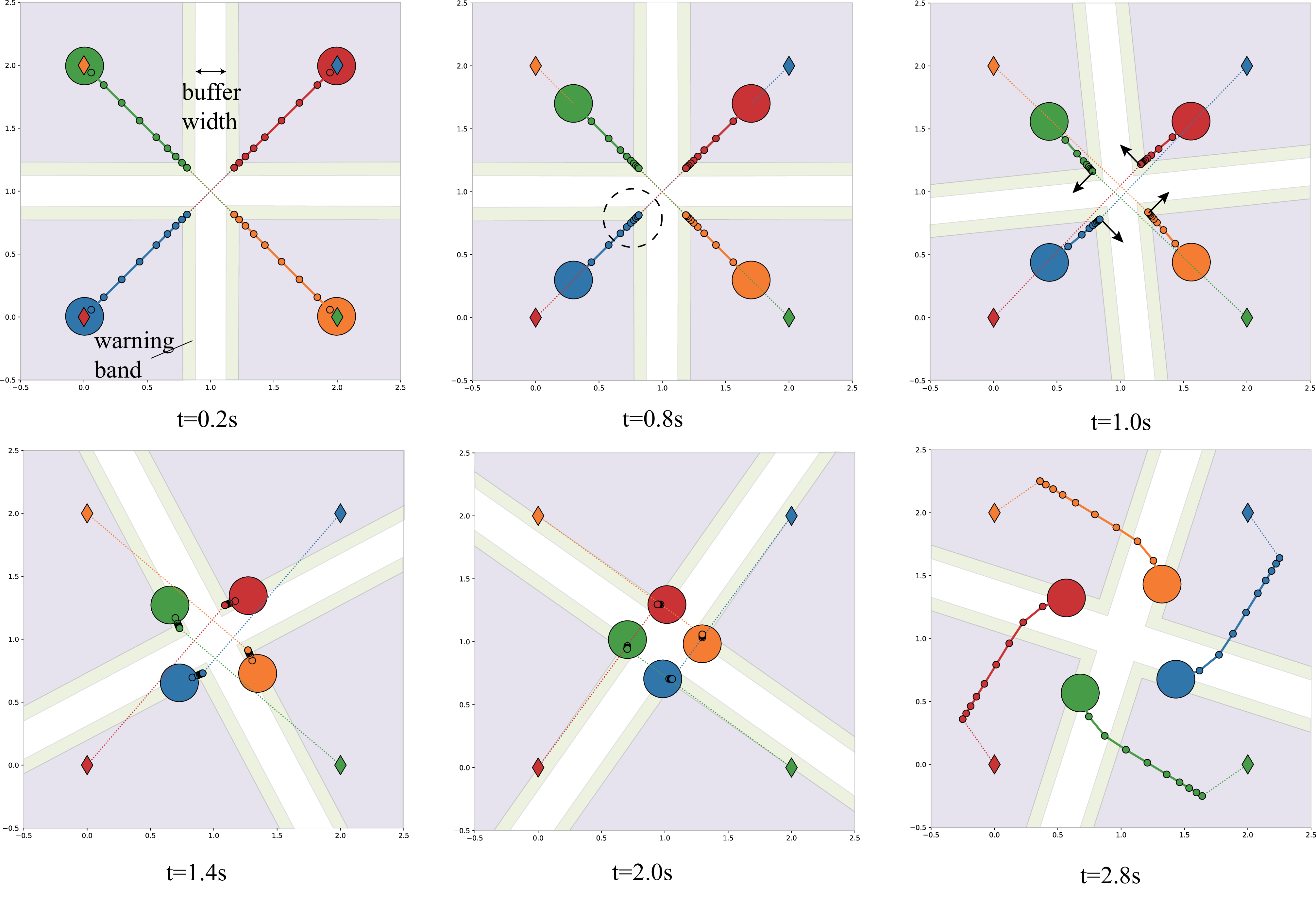}
	}
	\caption{ 
		The process of adopting the proposed right-hand rule where four robots want to swap their positions.
	}
	\label{2D_4_swap}
\end{figure}

\begin{figure}
	\centering
	\subfigure{
		\includegraphics[width=0.98\linewidth]{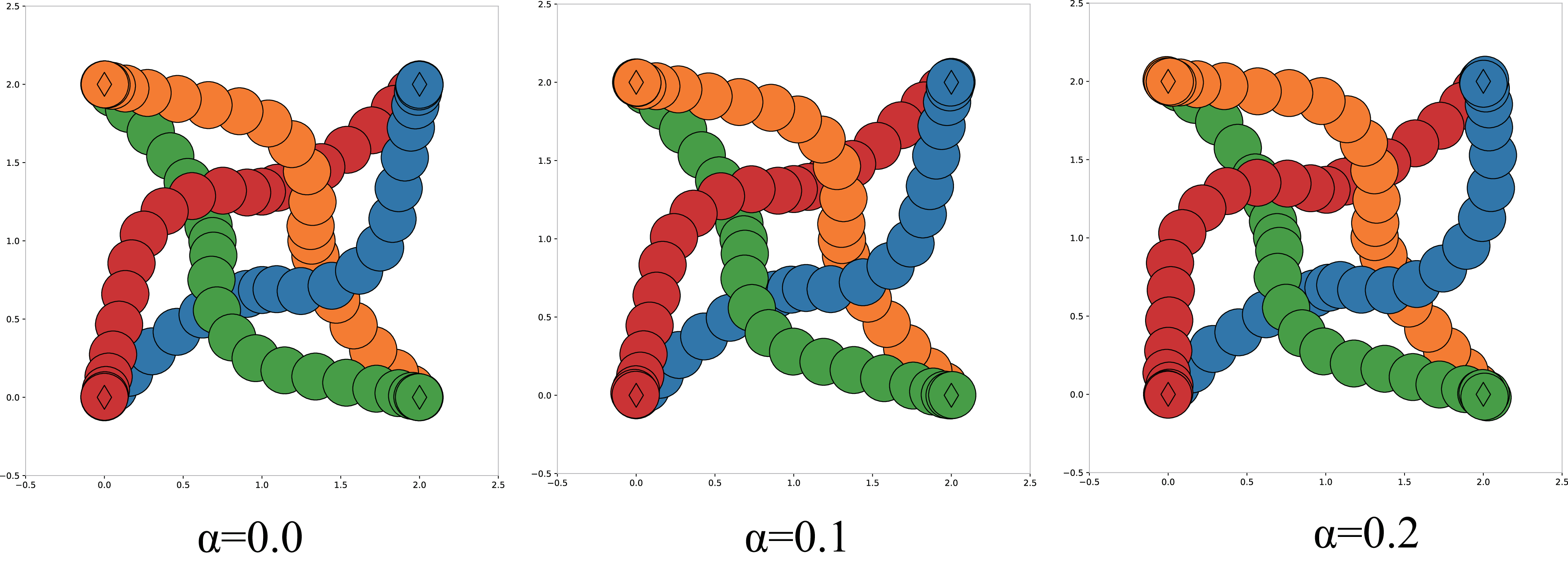}
	}
	\caption{
		Trajectories under different disturbances in a symmetric scenario.
	}
	\label{figure:2D_4_sensitivity}
\end{figure}

\section{Simulation and Experiment} \label{section:simulation-and-experiment}

In this section, the proposed algorithm is validated via numerical simulations and hardware experiments of large-scale multi-robot systems.
The algorithms are implemented in Python3, and publicly available at~\href{https://github.com/PKU-MACDLab/IMPC-DR}{https://github.com/PKU-MACDLab/IMPC-DR}.
Simulation and experiment videos can be found at~\href{https://youtu.be/IDcXoEer068}{https://youtu.be/IDcXoEer068}.
The convex optimizations are formulated by CVXPY~\cite{cvxpy} and solved by MOSEK~\cite{mosek}.
The numerical simulations include
some typical scenarios such as symmetric scenarios, narrow passage, position swapping,
and random transitions.
For the performance evaluation, our method IMPC-DR is compared with three other state-of-the-art methods:
iSCP \cite{Chen2015}, DMPC \cite{Luis2019} and BVC~\cite{Zhou2017}.
The implementation of iSCP and DMPC is based on~\cite{LuisCode}.

\subsection{Typical Scenarios} \label{subection:typical-simulations}

To begin with, some typical scenarios in MATG are considered first.
To be consistent with the model of UAVs in the experiments, the maximum velocity is set to~$v_\text{max}=1.0 {\rm m/s}$ and the maximum acceleration $a_\text{max}=1.5 {\rm m/s^2}$.
The sampling time~$h$ is chosen as $0.2$s and the horizon length is set to $K=10$ for all scenarios.
The minimum inter-robot distance  $r_\text{min}$ is chosen as $0.3$m and the width of warning band $\epsilon$ is set as $0.1$m.
In addition, the position penalty is set to~$Q_K=30.0$, $\rho_{0}=2.0$ and $\eta_0=2.0$.

\subsubsection{Symmetric Scenarios}

Symmetric scenarios are designed such that the initial and target positions of all robots are symmetric, which is the most common cause for deadlocks \cite{Grover2020}.
As shown in Fig.~\ref{2D_4_swap}, four robots located in a $2{\rm m} \times 2{\rm m}$ square transit to their antipodal positions.
The robots approach the center point initially at $t=0.2$s and the terminal positions have entered the warning band.
Then at $t=0.8$s, the condition for a terminal overlap holds as the terminal planned  positions overlap within the warning band and consequently the proposed resolution scheme is activated.
In other words, potential deadlocks are detected at this time instance, before they might happen.
As a result, from time $t=1.0$s, the robots begin to approach to their right-hand sides, as illustrated in Fig.~\ref{2D_4_swap}.
This adaptive process continues at $t=1.4$s, $2.0$s until the terminal positions leave the warning band and robots have escaped from the deadlock at time $t=2.8$s.
It is worth noting that without this resolution scheme, a deadlock eventually happens due to the symmetric configuration.

Furthermore, to validate the robustness of our deadlock resolution scheme,
artificial forces are applied to all robots as external disturbances, e.g., to mimic wind gust.
The forces follow a uniform Gaussian distribution with mean zero and standard deviation~$\sigma$.
As shown in Fig.~\ref{figure:2D_4_sensitivity}, when~$\alpha=\sigma/a_\text{max}$ is changed from~$0$ to $0.2$ with the symmetric scenario described above,
the proposed navigation and deadlock resolution scheme achieves the same level of deadlock resolution and similar completion time.

\begin{figure}[t!]
	\centering
	\subfigure{
		\includegraphics[width=1.0\linewidth]{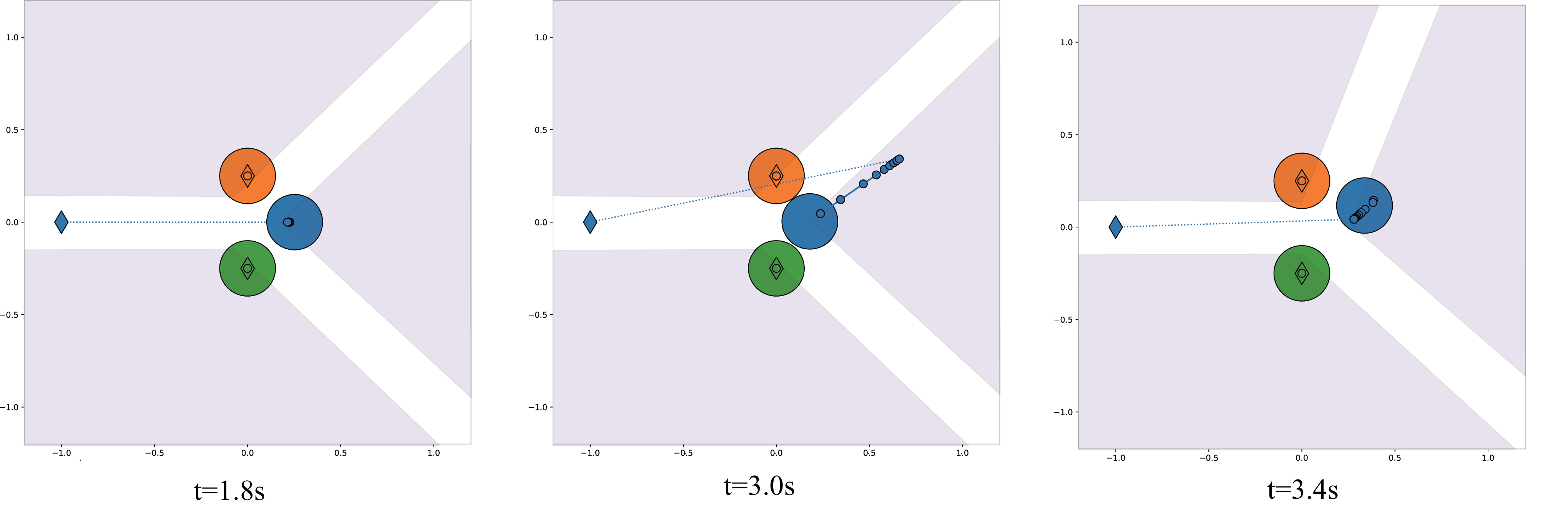}
		\label{2D_3_BVC-DR}
	}
	\quad
	\subfigure{
		\includegraphics[width=0.98\linewidth]{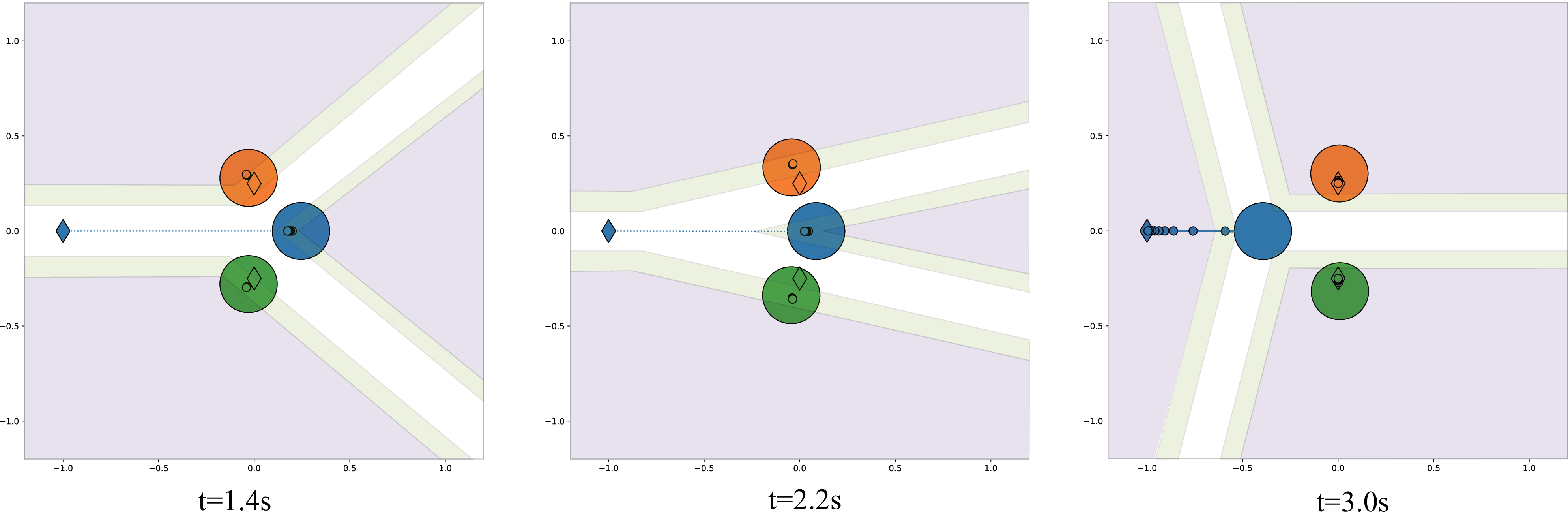}
		\label{2D_3_IMPC-DR}}
	\caption{ 
		The typical scenario of narrow passage where one robot passes through other two robots at target positions. 
		\textbf{Top}: the traditional right-hand rule suffers from livelock problem. 
		\textbf{Bottom}:  the proposed method enables the robots on the two sides to temporarily move away and leave space for the middle robot to pass through.
	}
	\label{2D_3}
\end{figure}

\subsubsection{Narrow Passage}
The second typical scenario is where a robot needs to pass through another two robots that are already at their respective target positions.
This scenario emphasizes the necessity of adding the warning band to the objective function of optimization~\eqref{eq:convex-program}.
As shown in Fig.~\ref{2D_3}, via the proposed algorithm,
when the robot in the middle approaches the other two robots, these two robots slowly move away from its target position to leave enough space for the middle robot to pass.
Concretely, when the middle robot approach to intersection position, another two robots are compelled to enter their warning band related to this robot as shown at $t=1.4$s.
Intuitively, the penalty term~\eqref{eq:C^i_w} in the objective function~\eqref{eq:C^i} can be decreased when both robots on the side move away from their targets until they are not located within the warning band anymore at around $t=3.0$s.
In contrast, the heuristic approach of choosing a detour point during deadlock suffers from the livelock problem.
Specifically, at time $t=3.0$s, the robot in the middle chooses a temporary target position and moves away once the deadlock is detected as all robots are static.
However, at around $t=3.4$s, it judges that it has escaped from this deadlock, and the target position is reset to the actual one.
Consequently, it comes back to the same deadlock equilibrium and this process repeats indefinitely.
Similar phenomenons can be found if artificial right-hand perturbations~\cite{Wang2017} are used as the heuristic method.

\begin{figure}[t]
	\centering
	\subfigure{
		\includegraphics[width=1.0\linewidth]{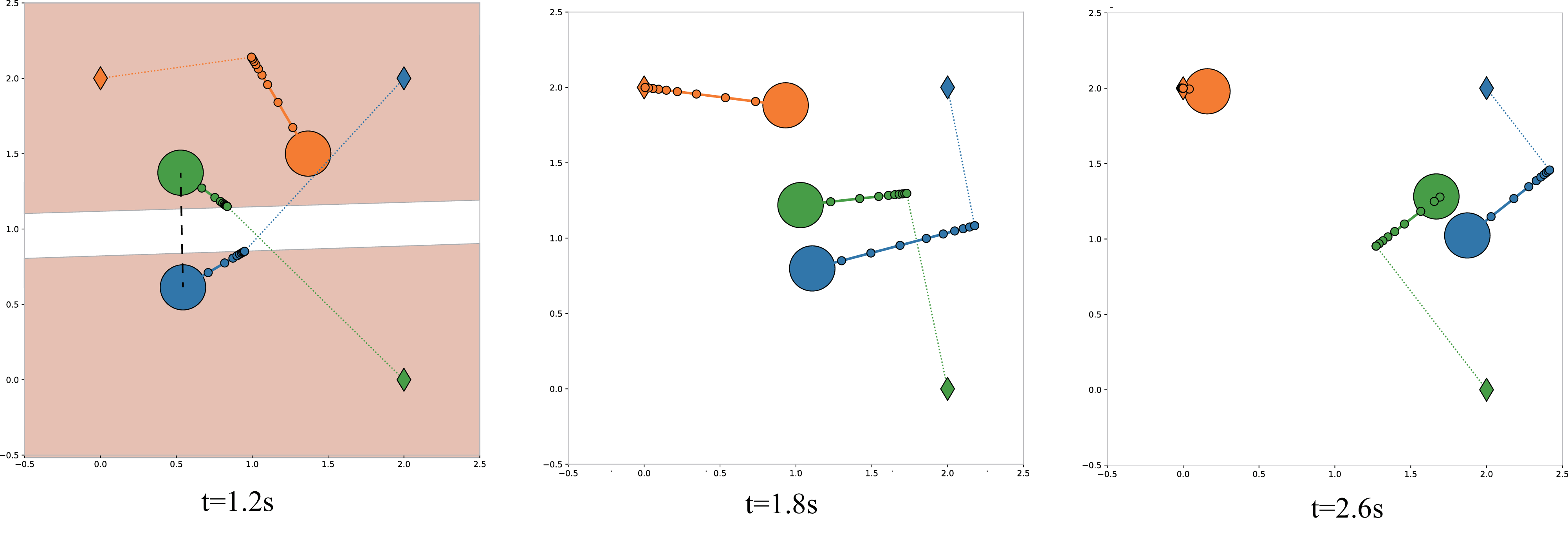}
		\label{2D_3_swap_BVC}
	}
	\quad
	\subfigure{
		\includegraphics[width=1.0\linewidth]{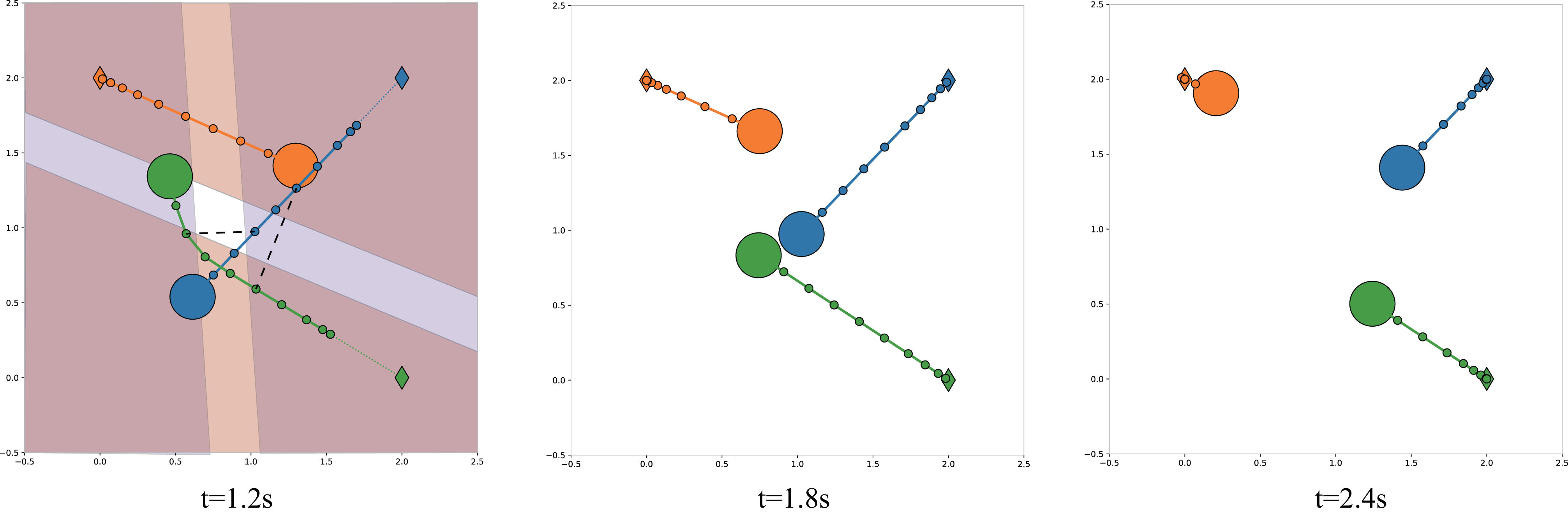}
		\label{2D_3_swap_IMPC-DR}
	}
	\caption{
		The typical scenario of swapping positions. 
		\textbf{Top}: The traditional BVC method may generate zigzag motions as BVC partitions the workspace only based on the current positions of all robots, as illustrated at $t=1.2s$. 
		\textbf{Bottom}: The proposed MBVC-WB however divides the workspace based on all future planned trajectories as shown at $t=1.2s$. Consequently, much smoother trajectories are generated.
	}
	\label{2D_3_swap}
\end{figure}

\subsubsection{Position Swapping}

The last scenario is where the robots swap their positions, as shown in Fig.~\ref{2D_3_swap}.
This scenario is designed to emphasize that the modified space partition constraint in~\eqref{eq: a p > b} leads to a more accurate separation among the robots and thus a higher utility rate of the workspace.
A comparison between the proposed method and the traditional BVC~\cite{Zhou2017} is shown in Fig.~\ref{2D_3_swap} for a particular setup.
Specifically, since the BVC only considers the current positions of all robots for space partition, all future positions in the planned trajectory are limited to this partition.
Thus, it often leads to an overly conservative navigation structure with excessive breaking and low efficiency.
In contrast, the proposed method enforces multiple space partitions based on the planned positions at each time step of horizon, i.e.,
the planned trajectories can be extended as how other robots plan to move.
This yields a much smoother and significantly more efficient navigation strategy.
This difference is apparent in Fig.~\ref{2D_3_swap}, where the robots accomplish the navigation task via the proposed method at~$t=3.0$s while it takes~$4.0$s for the traditional BVC method.
More detailed comparisons w.r.t. efficiency and success rate can be found in the next part.

\subsection{Random Transitions}

To systematically compare our method with other baselines including the incremental sequential convex programming (iSCP)~\cite{Chen2015}, the distributed model predictive control (DMPC)~\cite{Luis2019} and the buffered Voronoi cell (BVC)~\cite{Zhou2017},
the scenario of random transitions is designed,
where the initial and target positions are randomly chosen
in crowded 2D workspace and high-speed 3D space.
In total, $100$ random tests are generated for each case.
In each test, a navigation task is successful if all robots arrive at their target positions within considerable long time $50.0$s.
Note that for optimization-based methods, if the underlying optimization is infeasible and thus no solution exists, then the task fails.

\begin{figure}[t!]
	\centering
	\subfigure{
		\includegraphics[width=0.98\linewidth]{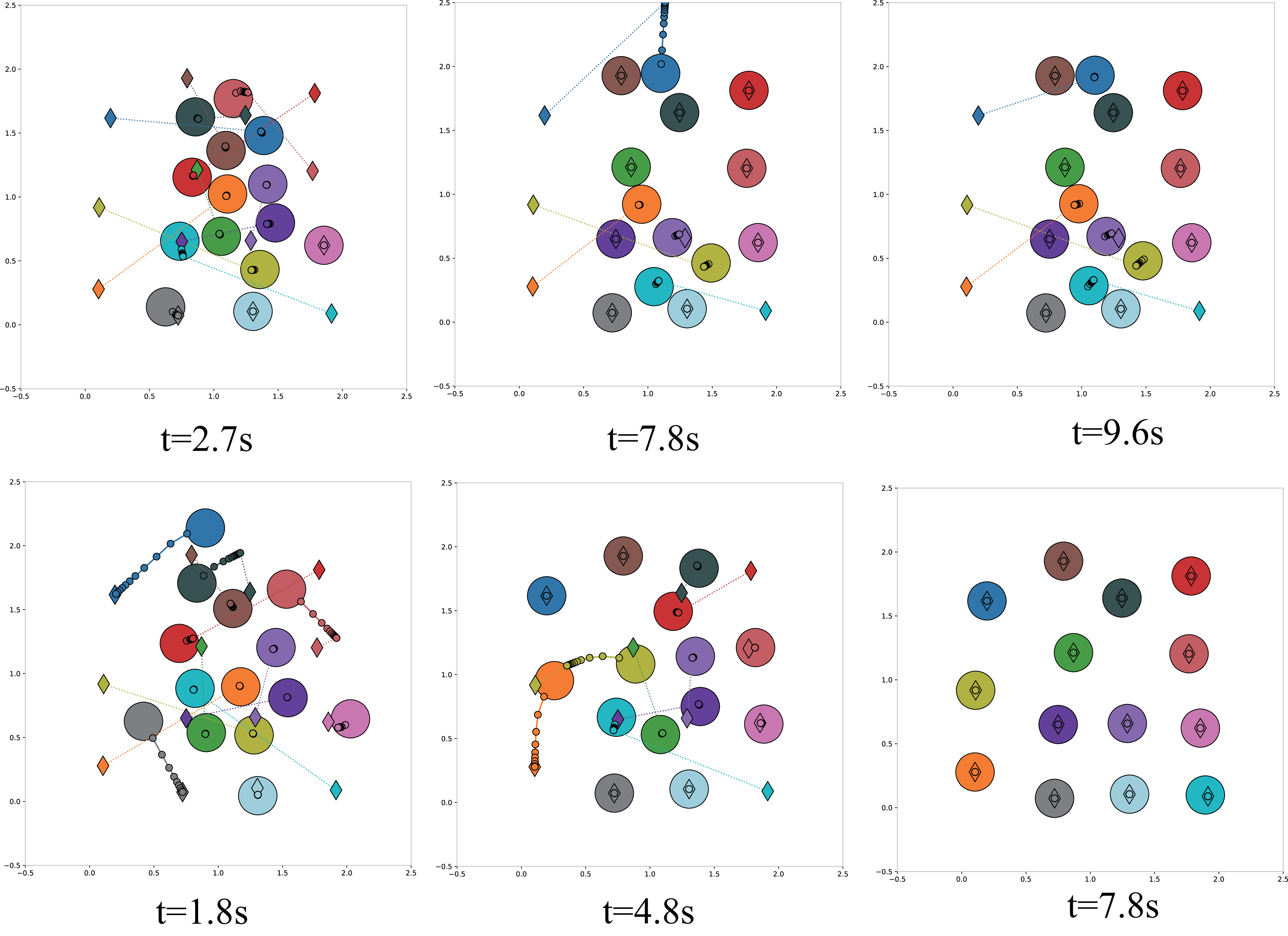}
	}
	\caption{Snapshots of the random transition in crowed navigation tasks in 2D by $14$ robots.
          \textbf{Top}: The traditional right-hand rule that chooses a detour point suffers from the livelock problem, where $4$ robots oscillate around the deadlock positions from time $t=7.8$s
          \textbf{Bottom}: The proposed method is applied to the same scenario, where potential deadlocks are resolved easily.}
	\label{figure:2D_14}
\end{figure}

\subsubsection{Crowded 2D Workspace}

The 2D workspace is set to~$2 \rm{m} \times 2$m and the number of robots ranges from~$2$ to~$14$.
The safety diameter, maximum velocity, maximum acceleration, the warning band width and other parameters are selected the same as before.
The only difference is that the sampling time~$h=0.15$s and the horizon length is set to $K=12$.
An example of $14$ robots is shown in Fig.~\ref{figure:2D_14}.
The results in terms of success rate and infeasibility are summarized in Table~\ref{table:2D crowded},
which shows that IMPC-DR achieves $100\%$ success rate and does not suffer from infeasibility at all.
Especially for the highly crowed case of $14$ robots, the success rate is much higher than other baselines.
More specifically, BVC~\cite{Zhou2017} with simple right-hand heuristic often suffers from the livelock problem described earlier in Section~\ref{section:analysis} with a success rate only at~$41\rm{\%}$,
of which an example is shown in Fig.~\ref{figure:2D_14}.
It is worth noting that when $\epsilon$ is set to~$0.1$m, $14$ robots within the finite space of~$2\rm{m} \times 2$m is almost at the \emph{highest} capacity.
One thousand random tests are conducted with a success rate of $100\%$ and zero deadlocks as well as no livelock.
This implies that the proposed method in this work has a good performance of deadlock resolution and livelock avoidance.
Furthermore, the completion time is evaluated for BVC \cite{Zhou2017} and the proposed method to illustrate the efficiency of MBVC-WB.
As provided in Table~\ref{table:transition time}, the proposed method has a significant decrease in transition time especially in a more crowded scenario.

\begin{table} [t!]
	\caption{Random transitions in crowded 2D scenarios}
	\label{table:2D crowded}
	\begin{tabular}{lllllllll}
		\toprule
		\multirow{2}{*}{Metric} & \multirow{2}{*}{Method} & \multicolumn{7}{c}{Number of robots}\\
		\cmidrule(lr){3-9}
		&&2 &4 &6 &8 &10 &12 &14 \\
		\midrule
		\multirow{5}{*}{Success} & IMPC-DR &\textbf{100} &\textbf{100} &\textbf{100} &\textbf{100} &\textbf{100} &\textbf{100} &\textbf{100}\\
		&BVC \cite{Zhou2017} &100 &100 &97 &94 &83 &65 &41\\
		&iSCP \cite{Chen2015} &100 &99 &94 &81 &58 &39 &12\\
		&DMPC \cite{Luis2019} &100 &99 &98 &91 &95 &73 &63\\
		\cmidrule(lr){1-9}
		\multirow{5}{*}{Infeas.} & IMPC-DR &\textbf{0} &\textbf{0} &\textbf{0} &\textbf{0} &\textbf{0} &\textbf{0} &\textbf{0}\\
		&BVC \cite{Zhou2017} &0 &0 &1 &0 &0 &2 &0\\
		&iSCP \cite{Chen2015} &0 &1 &6 &19 &20 &61 &88\\
		&DMPC \cite{Luis2019} &0 &1 &2 &9 &5 &27 &37\\
		\bottomrule
	\end{tabular}
\end{table}

\begin{table}[t!]
  \begin{center}
	\caption{time of random transitions in Crowded 2D space}
	\label{table:transition time}
	\begin{tabular}{llllllll}
		\toprule
		\multirow{2}{*}{Method} & \multicolumn{7}{c}{Completion Time $[s]$}\\
		\cmidrule(lr){2-8}
		&2 &4 &6 &8 &10 &12 &14 \\
		\midrule
		IMPC-DR             &1.98 &2.28 &2.72 &3.16 &4.31 &5.06 &6.20\\
		BVC \cite{Zhou2017} &2.01 &2.75 &3.26 &4.51 &6.21 &7.97 &10.85\\
		\bottomrule
	\end{tabular}
  \end{center}
\end{table}

\subsubsection{High-Speed 3D Workspaces}

\begin{figure}[t!]
	\centering
	\subfigure{
		\includegraphics[width=1.0\linewidth]{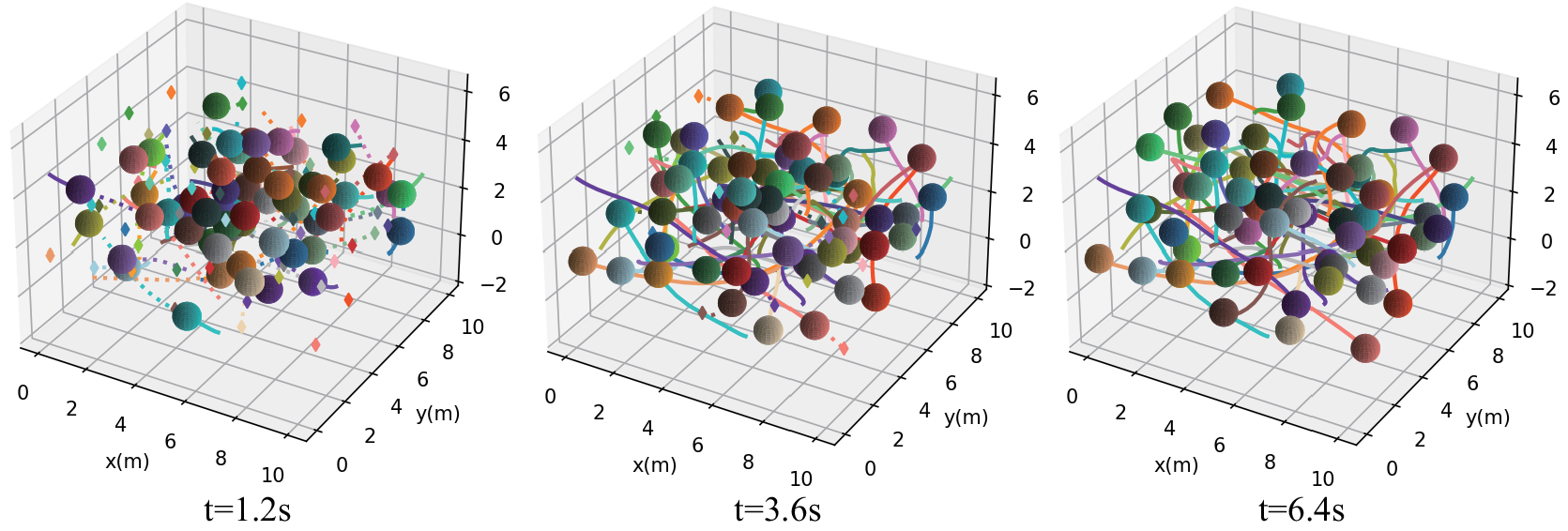}
	}
	\caption{
		Snapshots of high-speed random transitions within 3D via the proposed method by~$60$ robots.
	}
	\label{figure:rand_60}
\end{figure}

In this case, the maximum velocity and acceleration are increased to~$3$m/s and $2 \rm{m/s^2}$, respectively;
the 3D workspace is extended to $10 \rm{m} \times 10\rm{m} \times 5\rm{m}$.
The safety diameter of all robots is set to~$1.0$m in addition to the warning band width is extended to~$0.2$m.
As shown in Fig.~\ref{figure:rand_60}, $60$ robots can transit at a high speed with safety guarantee by IMPC-DR in this space.
Tests are performed for system sizes from $8$ to $60$,
of which the comparisons w.r.t. the success and infeasibility rate are summarized in Table~\ref{table:3D high}.
It can be seen that the performance remains almost the same as the 2D case in Table~\ref{table:2D crowded}.
In contrast, the performance of other baselines degraded \emph{significantly},
mostly due to the overaggressive trajectories and slow reaction to deadlocks.
This highlights the effectiveness of the proposed terminal constraints in \eqref{static constraint}.

\begin{table} [t]
	\caption{Random transitions in high-speed 3D scenarios}
	\label{table:3D high}
	\begin{tabular}{lllllllll}
		\toprule
		\multirow{2}{*}{Metric} & \multirow{2}{*}{Method} & \multicolumn{7}{c}{Number of robots}\\
		\cmidrule(lr){3-9}
		&&8 &16 &24 &32 &40 &50 &60 \\
		\midrule
		\multirow{5}{*}{Success} & IMPC-DR &\textbf{100} &\textbf{100} &\textbf{100} &\textbf{100} &\textbf{100} &\textbf{100} &\textbf{100}\\
		&BVC \cite{Zhou2017} &95 &58 &43  &35 &27 &21 &17\\
		&iSCP \cite{Chen2015} &49 &5 &0 &0 &0 &0 &0\\
		&DMPC \cite{Luis2019} &83 &36 &12 &0 &0 &0 &0\\
		\cmidrule(lr){1-9}
		\multirow{5}{*}{Infeas} & IMPC-DR &\textbf{0} &\textbf{0} &\textbf{0} &\textbf{0} &\textbf{0} &\textbf{0} &\textbf{0}\\
		&BVC \cite{Zhou2017} &5 &42 &57 &65 &73 &79 &83\\
		&iSCP \cite{Chen2015} &51 &95 &100 &100 &100 &100 &100\\
		&DMPC \cite{Luis2019} &17 &64 &88 &100 &100 &100 &100\\
		\bottomrule
	\end{tabular}
\end{table}

\begin{figure}[t]
	\centering
	\includegraphics[width=0.5\textwidth]{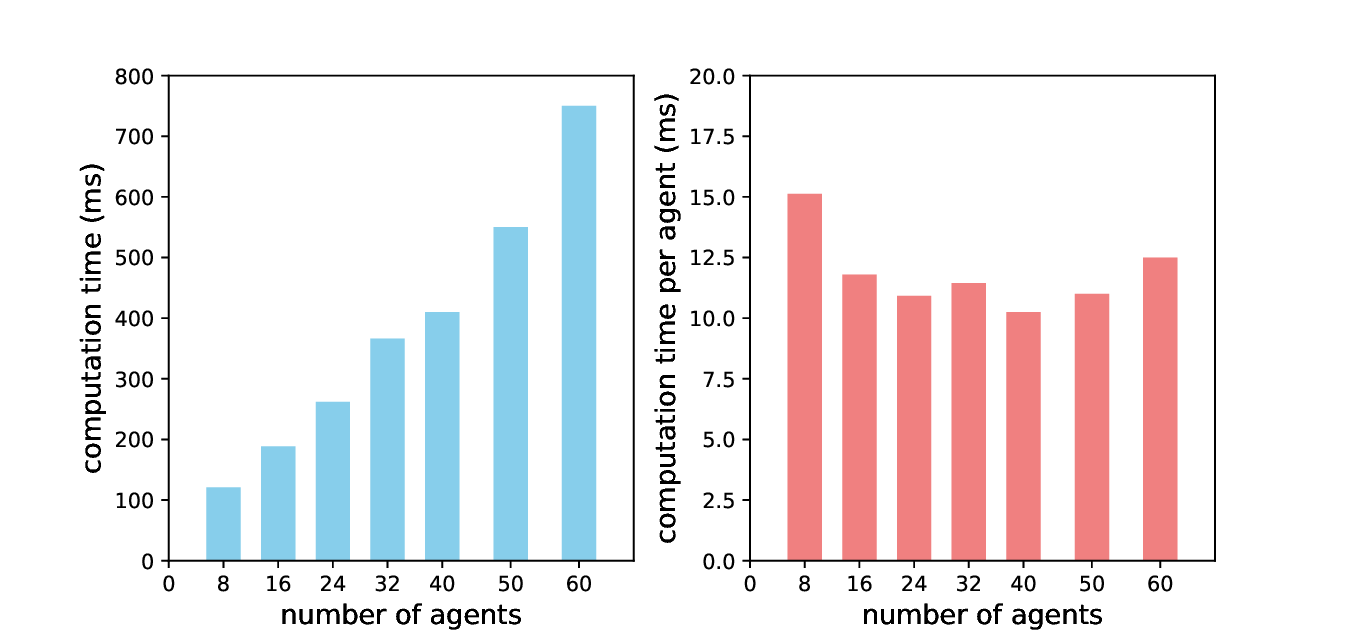}
	\caption{
		\textbf{Left}: Computation time over different system sizes.
        \textbf{Right}: Computation time per robot.
    }
	\label{figure:compuation-time}
\end{figure}

It is worth pointing out that for the above evaluation,
the computation of all robots is performed on one common computer.
However, the planning of each robot is an independent process.
Despite of being implemented in Python3 instead of C++,
it can still achieve a cycle time of~$0.75$s for the most complex case of~$60$ robots in~3D.
Fig.~\ref{figure:compuation-time} illustrates how the computation time changes as the number of robots increases.
It can be seen that the local computation time of each robot remains almost unchanged as the system size increases.
Thus, the proposed distributed method scales well with the number of robots.

\subsection{Experiments} \label{experiments}

To further validate the proposed method,
several experiments are performed on a nano quadrotor platform.

\begin{figure}[t]
	\centering
	\includegraphics[width=0.7\linewidth]{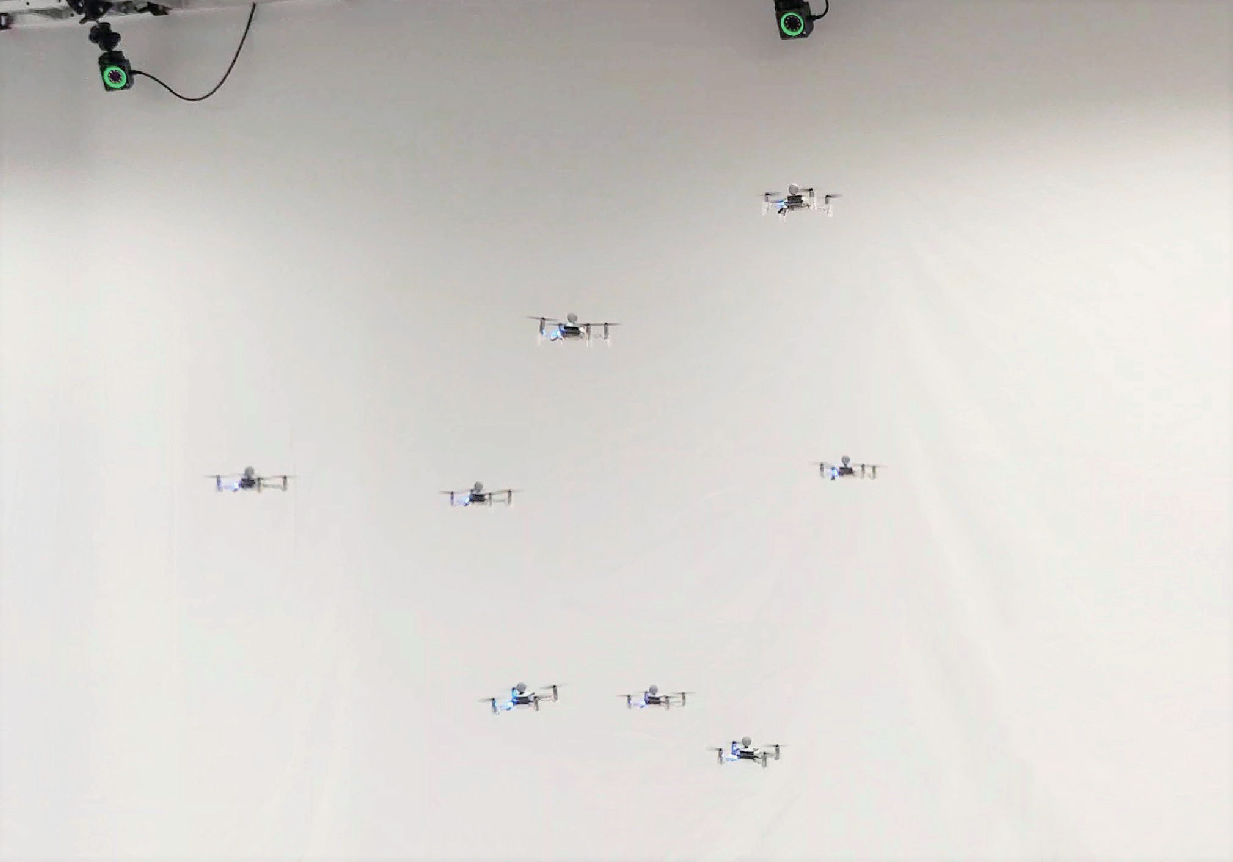}
		\caption{
			Hardware platform consists of a team of Crazyflies nano quadrotors, a motion capture system, and a control computer.
		}
	\label{real}
\end{figure}

\subsubsection{Hardware Setup}

As shown in Fig.~\ref{real}, the platform consists of several nano quadrotors named Bitcraze Crazyflie~2.1.
Their states in the workspace are captured by an indoor motion capture system \emph{OptiTrack}, of which the update frequency is $120$Hz. This information is sent to the main control computer where the proposed trajectory generation algorithm is carried out for all quadrotors.
The trajectory of quadrotor is fitted to a $7_{\text{th}}$-order polynomial and then sent to other quadrotors along with its state information
via high-frequency radio.
After receiving the position information, Kalman filter is used to estimate the current velocity and position. A feedback controller based on~\cite{Mellinger2011} is used to track the updated trajectory.

Furthermore, to avoid the inter-quadrotor air turbulence, the minimum distance between quadrotors $r_{\text{min}}$ is chosen as $0.3$m and the width of warning band $\epsilon$ is chosen as $0.1$m. 
The maximum velocity and acceleration of Crazyflies are set to~$1\rm{m/s}$ and $1\rm{m/s^2}$, respectively, to ensure safety. Lastly, the sampling time~$h$ is set to $0.2$s and the horizon length to~$15$, to balance the control performance and the computation burden.

\begin{figure}[htbp]
	\centering
	\subfigure[Twenty robots transit to their antipodal positions in a circle. \textbf{Left}: Robot Trajectories. \textbf{Right}: Inter-robot distances.]{
		\includegraphics[width=0.18\textwidth]{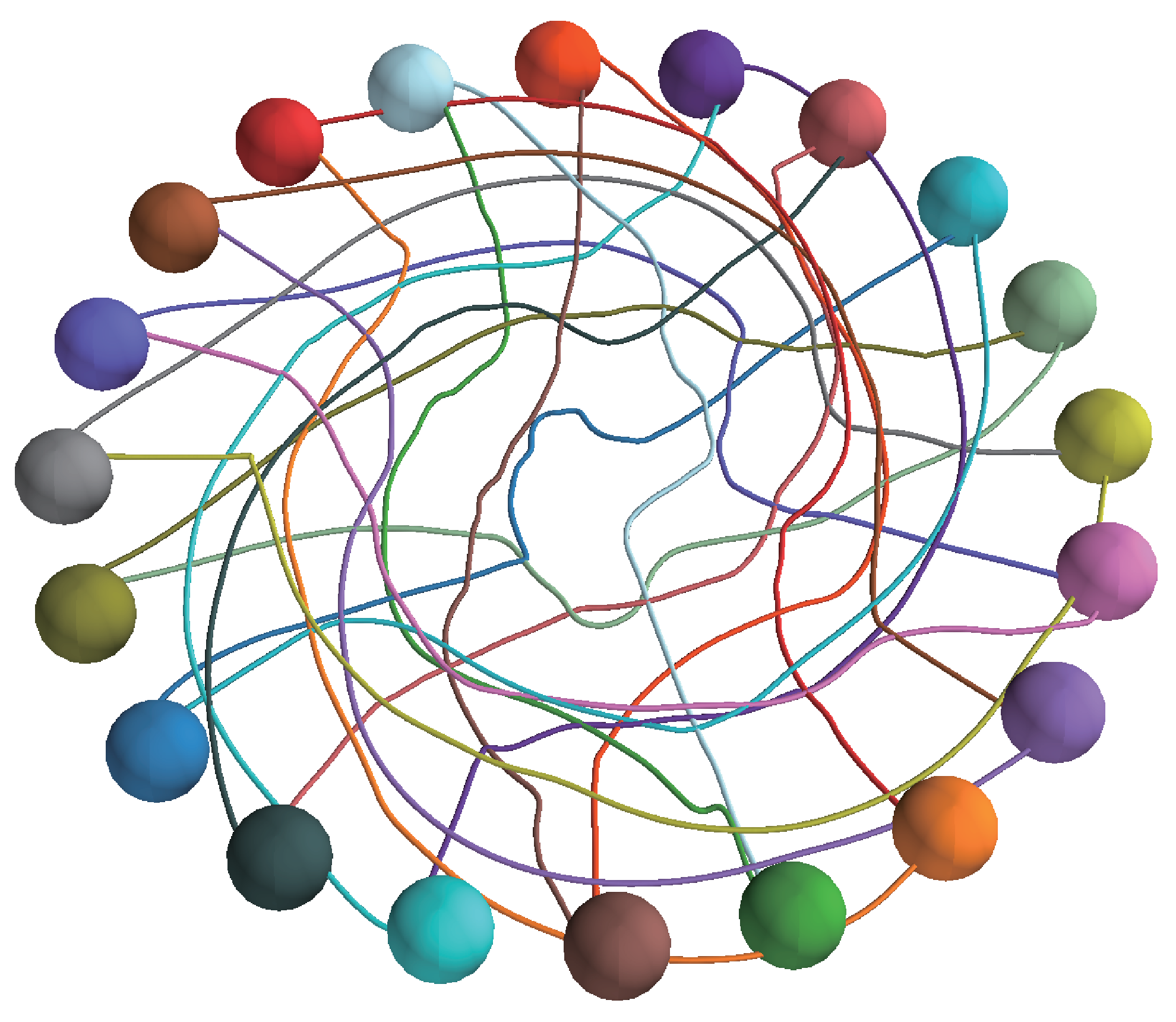}
		\includegraphics[width=0.22\textwidth,height=0.20\textwidth]{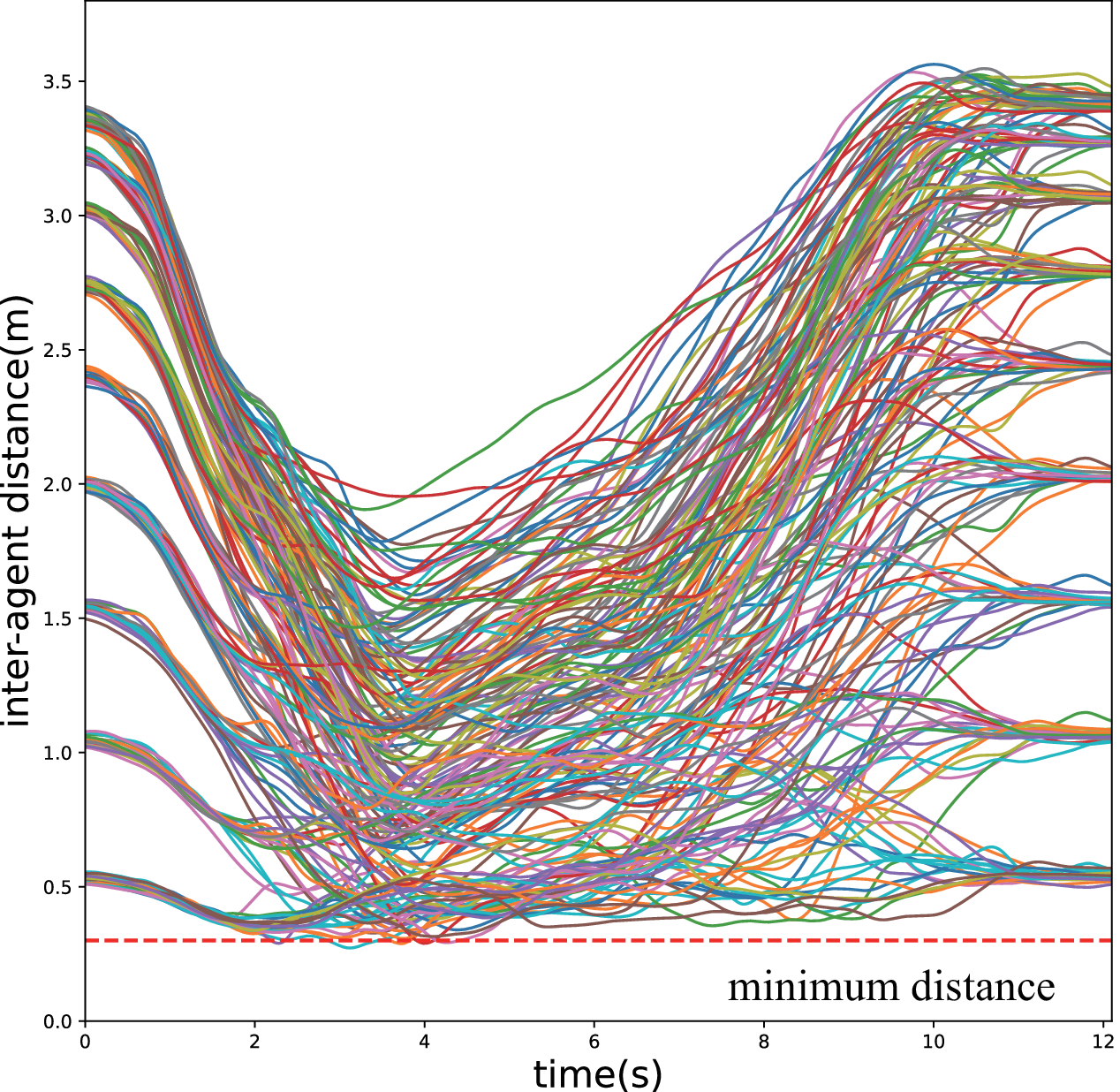}
		\label{2D-20-swap-real}
	}
	\quad
	\subfigure[8 crazyfies in a cube fly to their antipodal positions. \textbf{Left}: Robot Trajectories. \textbf{Right}: Inter-robot distances.]{
		\includegraphics[width=0.19\textwidth]{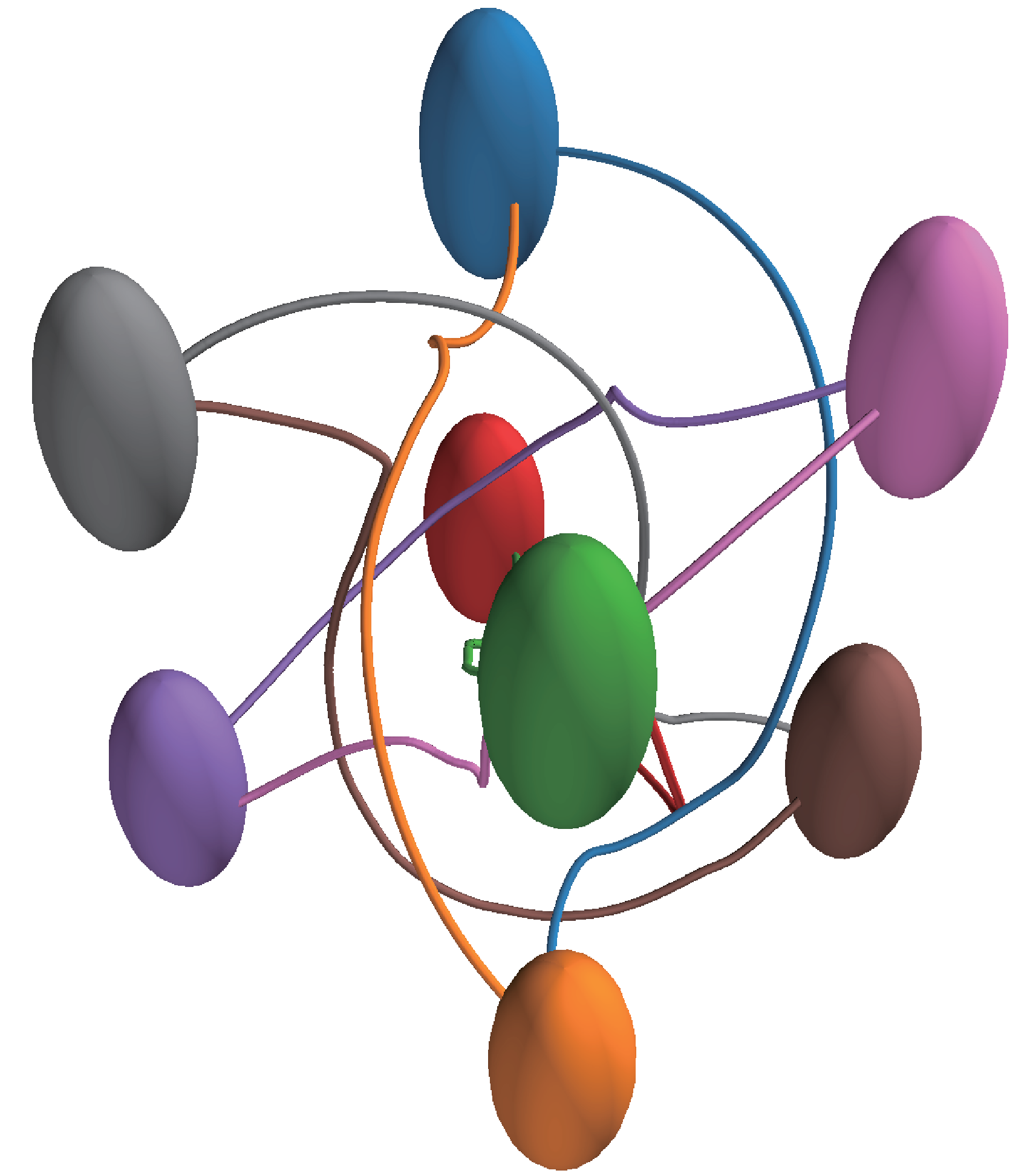}
		\includegraphics[width=0.22\textwidth]{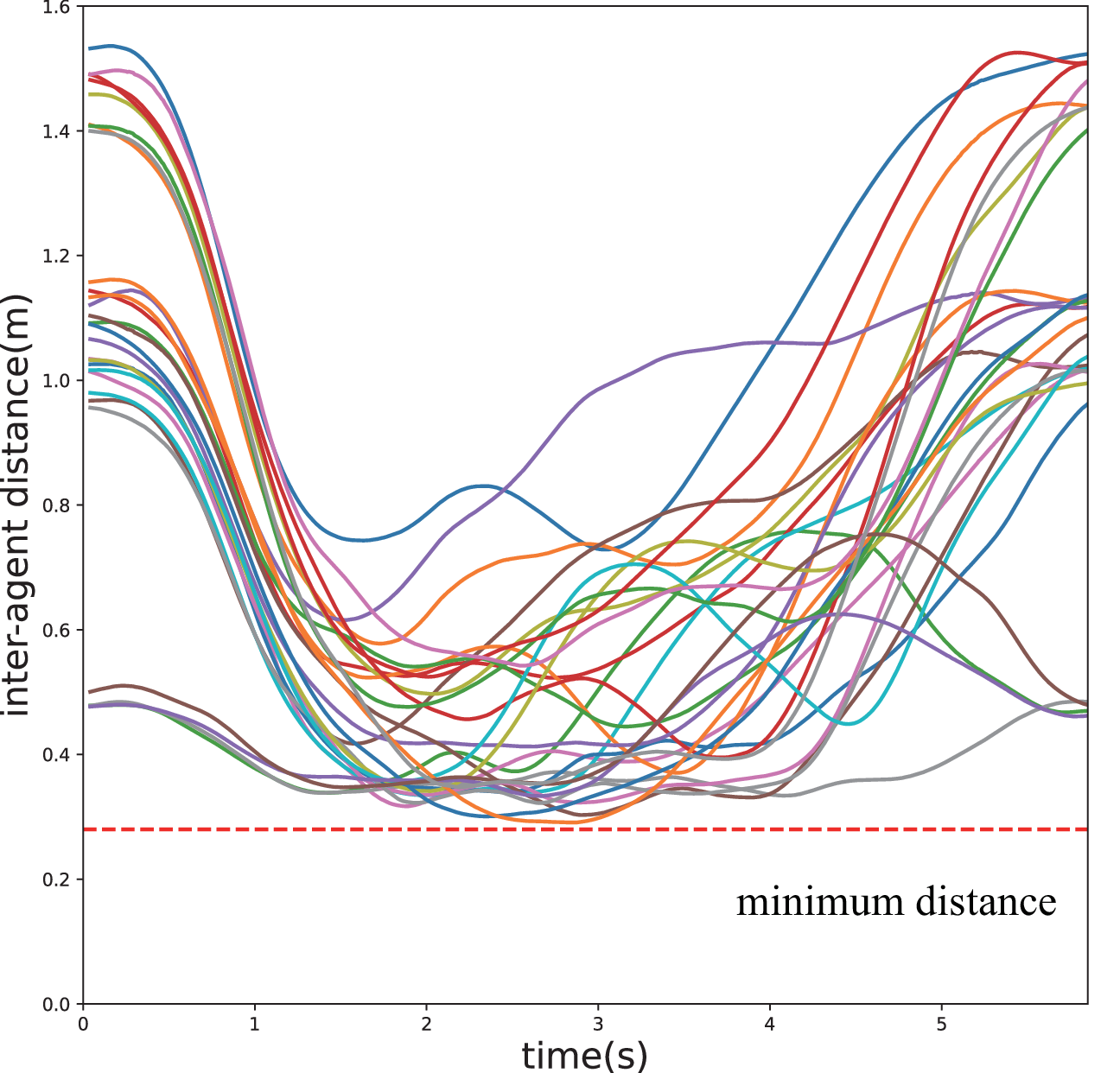}
		\label{figure:8-cube}
	}
	\caption{Hardware experiments of several scenarios, which can be compared with the numerical simulations in Section~\ref{subection:typical-simulations}.}
\end{figure}

\subsubsection{Results}

The first experiment is the symmetry scenario where $20$ quadrotors transit to their antipodal position in a 2D circle with radius $1.7$m.
As shown in Fig.~\ref{2D-20-swap-real}, the navigation task is accomplished within~$11.3$s with smooth and safe trajectories.
Due to the symmetric setup, each agent is potentially blocked by its left and right neighbors at about $t=2.0$s.
Via the proposed resolution scheme, the repulsive force from the left hand is increased and thus larger than the right-hand force, yielding a right-hand rotation.
This pattern continues for about $7$s such that most of the underlying agents are close to their targets.
A slight violation of the safety constraint (less than $0.03$m) happens around $2.6$s due to the strong air turbulence in the confined space, which however can be compensated by enlarging the safety margin, as did in \cite{Luis2019,Park2021}. In the second experiment, $8$ quadrotors transit to their antipodal corners in a cube of $1\rm{m^3}$.
To consider the downwash effect caused by flying quadrotors,
the robots are encapsulated within an ellipsoid of diameter $0.3$m in $xy$-axis and $0.7$m in $z$-axis.
The proposed IMPC-DR formulation in~\eqref{eq:convex-program} can be easily adapted
to this configuration.
The final trajectories are shown in Fig.~\ref{figure:8-cube}, where the right-hand rotation is also visible when the robots avoid each other.

\section{Conclusion} \label{section:conclusion}

This work has proposed a novel and effective navigation algorithm for multi-robot systems.
Its core part is called infinite-horizon model predictive control with deadlock resolution.
Different from many heuristic methods, it can effectively detect and resolve deadlocks online before they even happen.
In addition, it provably ensures recursive feasibility of the underlying optimization at each replanning.
It is a fully distributed method that requires only local communication and scales well with the number of robots.
Compared with other state-of-the-art baselines, its advantages especially in crowded and high-speed scenarios are significant, as demonstrated both in simulations and hardware experiments.
Future work includes extensions to the obstacle-rich environment.

\appendices

\section{Lemma~\ref{lemma:distance}} \label{appendix:lemma-distance}

\begin{lemma} \label{lemma:distance}
	Consider two line segments: the line segment from $p_1$ to $p_2$ and the line segment from $q_1$ to $q_2$, where $p_1,p_2,q_1,q_2\in \mathbb{R}^d$, $d=2,3$.
	If
	\begin{equation} \label{condition}
		\|r_{1}\|,\,\|r_{2}\| \geq \sqrt{r_{\min}^{2} + \frac{1}{4} \|l_{2}-l_{1}\|^2}
	\end{equation}
	is satisfied for $r_{1}=q_{1}-p_{1}$,
	$r_{2}=q_{2}-p_{2}$,
	$l_{1}=p_{2}-p_{1}$ and
	$l_{2}=q_{2}-q_{1}$,
	then
	\begin{equation} \label{equa}		\|p_{1}+t\left(p_{2}-p_{1}\right)-q_{1}-t\left(q_{2}-q_{1}\right)\| \geq r_{\min }
	\end{equation}
	holds, $\forall t \in[0,1]$.
\end{lemma}

\begin{proof} [Proof of Lemma~\ref{lemma:distance}]
	It is trivial to show that the left-hand side of~\eqref{equa} is equivalent to
	\begin{equation*}
		p_{1} + t \left( p_{2} - p_{1} \right) - q_{1} - t \left( q_{2} - q_{1} \right)  \\
		= r_{1}+t\left(r_{2}-r_{1}\right).
	\end{equation*}
	Hence, it suffices to prove that
	\begin{equation*}
		\|r_{1}+t\left(r_{2}-r_{1}\right)\| \geq r_{\min }, \ \forall t \in[0,1].
	\end{equation*}

	Introduce a function~$F(t)=\|r_{1}+t\left(r_{2}-r_{1}\right)\|^2$, $t\in [0,1]$.
	Moreover, set $F(t_{\text{min}})=\min_{t \in [0,1]} F(t)$.
	Consider the following three cases:

	(a) $t_{\text{min}}=0$: $F(0)=r_{1}^\mathrm{T} r_{1} \geq r_{\min }^{2}$.

	(b) $t_{\text{min}}=1$: $F(1)=r_{2}^\mathrm{T} r_{2} \geq r_{\text{min}}^{2}$.

	(c) $0<t_{\text{min}}<1$: $F(t_{\text{min}})$ is given by
	\begin{equation*}
		F(t_{\text{min}})=\frac{r_{1}^\mathrm{T} r_{1} r_{2}^\mathrm{T} r_{2}-r_{1}^\mathrm{T} r_{2} r_{1}^\mathrm{T} 	r_{2}}{\left(r_{2}-r_{1}\right)^\mathrm{T}\left(r_{2}-r_{1}\right)},
	\end{equation*}
	where
	\begin{equation}\label{t_min}
		t_{\text{min}}=- \frac{r_{1}^\mathrm{T}(r_2-r_1)}{\left(r_{2}-r_{1}\right)^\mathrm{T}\left(r_{2}-r_{1}\right)}.
	\end{equation}
	Note that $F(t_{\text{min}}) \geq r_{\min }^{2}$ is equivalent to
	\begin{equation*}
		r_{1}^\mathrm{T} r_{1} r_{2}^\mathrm{T} r_{2}-r_{1}^\mathrm{T} r_{2} r_{1}^\mathrm{T} r_{2} \geq r_{\min 	}^{2}\left(r_{2}-r_{1}\right)^\mathrm{T}\left(r_{2}-r_{1}\right),
	\end{equation*}
	which can be rewritten as
	\begin{equation} \label{result2}
		\begin{aligned}
			& \left(r_{1}^\mathrm{T} r_{1}-r_{\min }^{2}\right)\left(r_{2}^\mathrm{T} r_{2}-r_{\min }^{2}\right) \\
			& \geq\left(r_{1}^\mathrm{T} r_{2}-r_{\min }^{2}\right)\left(r_{1}^\mathrm{T} r_{2}-r_{\min }^{2}\right).
		\end{aligned}
	\end{equation}
	Now, consider the following two cases:
	\begin{itemize}
		\item[(i)] $r_{1}^\mathrm{T} r_{2}-r_{{\text{min}} }^{2} \geq 0$.
		Since $t_{\text{min}} \in (0,1)$, it follows from \eqref{t_min} that~$r_{1}^\mathrm{T} r_{1} \geq r_{1}^\mathrm{T} r_{2}$ and $r_{2}^\mathrm{T} r_{2} \geq r_{1}^\mathrm{T} r_{2}$. Thus,
		\begin{equation*}
			\begin{aligned}
				r_{1}^\mathrm{T} r_{1}-r_{\min }^{2} \geq r_{1}^\mathrm{T} r_{2}-r_{\min }^{2} \geq 0, \\
				r_{2}^\mathrm{T} r_{2}-r_{\min }^{2} \geq r_{1}^\mathrm{T} r_{2}-r_{\min }^{2} \geq 0,
			\end{aligned}
		\end{equation*}
		which induces~\eqref{result2}.

		\item[(ii)] $r_{1}^\mathrm{T} r_{2}-r_{\min }^{2} < 0$.
		Without loss of generality, assume that~$r_{1}^\mathrm{T} r_{1} \leq r_{2}^\mathrm{T} r_{2}$.
		Then, it is easy to show that~$r_{1}^\mathrm{T} r_{1} \geq r_{\min }^{2}+\frac{1}{4}\|l_{2}-l_{1}\|^2$, given~\eqref{condition}.
		Combining with the simple fact that
		$l_{2}-l_{1}=q_{2}-q_{1}-p_{2}+p_{1}=r_{2}-r_{1}$ leads to
		\begin{equation*}
			\begin{aligned}
				r_{1}^\mathrm{T} r_{1} &\geq r_{\min }^{2}+\frac{1}{4}\|r_{2}-r_{1}\|^2 \\
				&= r_{\min }^{2}+\frac{1}{4}\left(r_{1}^\mathrm{T} r_{1}+r_{2}^\mathrm{T} r_{2}-2 r_{1}^\mathrm{T} r_{2}\right)  	\\
				&\geq  r_{\min }^{2}+\frac{1}{4}\left(2 r_{1}^\mathrm{T} r_{1}-2 r_{1}^\mathrm{T} r_{2}\right).
			\end{aligned}
		\end{equation*}

		After re-organizing the terms, we have
		\begin{equation*}
			r_{2}^\mathrm{T} r_{2} \geq r_{1}^\mathrm{T} r_{1} \geq 2 r_{\min }^{2}-r_{1}^\mathrm{T} r_{2},
		\end{equation*}
		and
		\begin{equation*}
			\begin{aligned}
				r_{1}^\mathrm{T} r_{1}-r_{\min }^{2} \geq r_{\min }^{2}-r_{1}^\mathrm{T} r_{2} \geq 0, \\
				r_{2}^\mathrm{T} r_{2}-r_{\min }^{2} \geq r_{\min }^{2}-r_{1}^\mathrm{T} r_{2} \geq 0,
			\end{aligned}
		\end{equation*}
		which implies \eqref{result2}.
	\end{itemize}

	Now, the proof is completed.
\end{proof}

\section{ Lemma~\ref{lemma:deadlock}} \label{appendix:deadlock}

\begin{lemma} \label{lemma:deadlock}
	Via the proposed navigation scheme, if two robots $1$ and $2$ are struck in deadlocks, the following properties hold:
	
	1)  $p^1$, $p^1_\text{target}$, $p^2$ and $p^2_\text{target}$ are collinear.
	
	2) the deadlocks shown in Fig.~\ref{figure:2_deadlock}(a) are unstable.
\end{lemma}

\begin{figure}[t]
	\centering
	\subfigure[The angle between $p^i_K(t-h)-p^i_\text{target}$ and $\tau$, represented by the green curve, is larger than $\frac{\pi}{2}$.]{
		\includegraphics[width=0.9\linewidth]{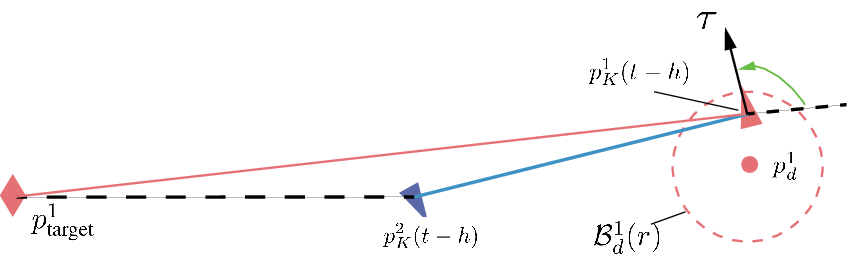}
		\label{figure:unstable_deadlock-1}
	}
	\quad
	\subfigure[The angle $\beta(t)$ has the tendency to be increased such that $p^1_K$ or $p^2_K$ cannot be bounded.]{
		\includegraphics[width=0.9\linewidth]{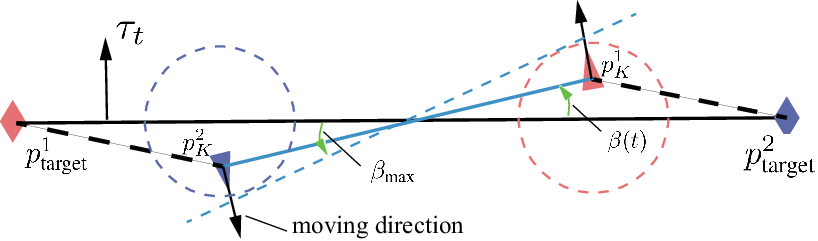}
		\label{figure:unstable_deadlock-2}
	}
	\caption{
		The deadlock shown in Fig.~\ref{figure:2_deadlock}(a) is unstable.
	}
\end{figure}

\begin{proof}
	Since it is proven in Theorem~\ref{theorem:deadlock-property} that the summed forces equal to zero, it implies that~$a^{12}_K=- \frac{p^i_\text{target} - p^1_K}{\|p^i_\text{target} - p^1_K\|} = - \frac{p^i_\text{target} - p^1}{\|p^i_\text{target} - p^1\|}$.
	Furthermore, via~\eqref{a-b-definition}, it holds that $a^{12}_K=\frac{ p^1_K-p^2_K } { \|p^1_K-p^2_K\| }=\frac{ p^1-p^2 } { \|p^1-p^2\| }$.
	Thus, $p^1_\text{target}$, $p^1$ and $p^2$ are collinear.
	Via similar analyses, $p^2_\text{target}$, $p^1$ and $p^2$ are also collinear.
	Consequently, $p^1$, $p^1_\text{target}$, $p^2$ and $p^2_\text{target}$ are all collinear.

	In the following, we will show that the deadlocks in Fig.~\ref{figure:2_deadlock}(a) are unstable. 
	As shown in Fig.~\ref{figure:unstable_deadlock-1}, define $\mathcal{B}^1_d(r_d) \triangleq \left\{ p \ | \ \|p-p^1\| \leq r_d \right\}$ as the ball of radius~$r_d$ around the deadlock position of robot~$1$.
	The radius~$r_d$ is chosen such that 1) if  $\|v^1\|<\frac{2 r_d}{h}$ holds, then $\|v^1_k\|<\frac{2 r_d}{h}$ and $\|u^1_k\|<\frac{4 r_d}{h^2}$ also hold, yielding that $\| \Theta_v v^1_k \| < v_{\text{max}}$  and~$\| \Theta_a u^1_{k-1} \| < a_{\text{max}}$ where $k \in \mathcal{K}$;
	2) if $p^i_k$, $p^i_k(t-h) \in \mathcal{B}^i_d(r_d)$ holds for $i=1,2$ and $k \in \mathcal{K}$, then the constraints ${a^{12}_K}^\mathrm{T} p^1_K = b_{K}^{12} + w^{12}$ and
	${a_{k}^{12}}^\mathrm{T} p_{k}^{1} > b_{k}^{12}$ hold for $k \in \tilde{\mathcal{K}}$.

	Thereafter, the motion tendency of the underlying robots are analyzed when they are bounded by $\mathcal{B}^1_d(r_d)$ and $\mathcal{B}^2_d(r_d)$.
	At time $t-h$, consider the vector
	\begin{equation*} \label{eq: r-definition}
		\begin{aligned}
		r_\tau = & \left( (p^2_K(t-h)-p^1_\text{target}) \times  (p^1_K(t-h) - p^2_K(t-h)) \right) \\
		& \times (p^1_K(t-h) - p^2_K(t-h)),
		\end{aligned}
	\end{equation*}
	along which the unit vector is defined as $\tau = \frac{r_\tau}{\|r_\tau\|}$.
	Consider the planning process at time~$t$,
    it can be shown that for robot $1$, the following holds
    \begin{equation} \label{eq: C1psi}
   		\begin{aligned}
		 & \tau^\mathrm{T} \frac{ \partial C^1 }{ \partial p^1_K } |_{p^1_K=p^1_K(t-h)}\\
		        &= \tau^\mathrm{T} \frac{ \partial C^1_p }{ \partial p^1_K } |_{p^1_K=p^1_K(t-h)} - \tau^\mathrm{T} \frac{ \partial C^1_w }{ \partial p^1_K }  |_{p^1_K=p^1_K(t-h)}\\
		    &= \tau^\mathrm{T} \frac{ \partial C^1_p }{ \partial p^1_K } |_{p^1_K=p^1_K(t-h)},
	    \end{aligned}
	\end{equation}
    where $ \tau^\mathrm{T} \frac{ \partial w^{12} }{ \partial p^1_K }
    = \tau^\mathrm{T} \frac{ \partial \left\{ {a^{12}_K}^\mathrm{T} p^1_K - b_{K}^{12} \right\} }{ \partial p^1_K }  
    = \tau^\mathrm{T} a^{12}_K =0$
    leads to
    $\tau^\mathrm{T} \frac{ \partial C^1_w }{ \partial p^1_K } = \frac{ \partial C^1_w }{ \partial w^{12} }  \tau^\mathrm{T} \frac{ \partial w^{12} }{ \partial p^1_K }=0$.
    Meanwhile, since both $\|\Theta_a u_{k-1}^{i}\| < u_{ \text{max} } $ and $\| \Theta_v v_{k}^{i}\| < v_{\text{max} }$ hold for $k \in \mathcal{K}$, \eqref{eq:KKT-p} and
    ${ }^{p} \nu_{k}^{1}=0$, $\forall k \in \tilde{\mathcal{K}}$ can be derived similarly as in the proof of Theorem~\ref{theorem:deadlock-property}.
    Furthermore, owing to ${a_{k}^{12}}^\mathrm{T} p_{k}^{1} > b_{k}^{12}$, it can be obtained that $\lambda_{k}^{12}=0$.
    Combining this with~\eqref{eq:KKT-p}, the equality
    $\frac{\partial \mathcal{L}^{1}}{\partial p_{k}^{1}}=\frac{\partial C^{1}}{\partial p_{k}^{1}}=0$ and
    $Q_k({p^1_k-p^1_{k+1}})+Q_{k-1}({p^1_k-p^1_{k-1}})=0$ hold, $k \in \tilde{\mathcal{K}}$.
    As a result, $p^1_k$, $k=0,1,\ldots,K$ are collinear, which induces
    $Q_k\|{p^1_k-p^1_{k+1}}\|=Q_{k-1}\|p^1_k-p^1_{k-1}\|$.
    Since it has been shown that $\sum_{k=0}^{K-1}\|p^1_k-p^1_{k+1}\|=\|{p^1_K-p^1_0}\|$, 
    it follows that
    $Q_k\|{p^1_k-p^1_{k+1}}\|=1/{\sum_{k=0}^{K-1} \frac{1}{Q_k}} \|{p^1_K-p^1_0}\|$.
    Since $\lim_{Q_0 \rightarrow 0^+} {1/{\sum_{k=0}^{K-1}} \frac{1}{Q_k}} = 0$, it further implies that
    $\|Q_k \left(p^1_K-p^1_{K-1}\right)\|=0$ and $Q_k ({p^1_{k+1}-p^1_k})=0$ hold for $k = 0,1,\ldots,K-1$.
    Then, since $Q_k>0$ exists for $k \in \tilde{\mathcal{K}}$, it is clear that $p^1_{k+1}=p^1_k$ holds for $k \in \tilde{\mathcal{K}}$.
    According to $p^1_K=p^1_{K-1}$, it can be derived that
    $$\begin{aligned}
    \frac{ \partial C^1_p }{ \partial p^1_K } &= Q_{K-1} ({p^1_K-p^1_{K-1}}) + {Q_K (p^1_K-p^1_\text{target})} \\&= Q_K (p^1_K-p^1_\text{target}).
    	    \end{aligned}$$
    Along with~\eqref{eq: C1psi}, the following relation holds
    \begin{equation*}
    	\tau^\mathrm{T} \frac{ \partial C^1 }{ \partial p^1_K } |_{p^1_K=p^1_K(t-h)}
    	= Q_K \tau^\mathrm{T} (p^1_K(t-h)-p^1_\text{target}).
    \end{equation*}
	As depicted in Fig.~\ref{figure:unstable_deadlock-1}, it is evident that the angle between $p^1_K(t-h)-p^1_\text{target}$ and $\tau$ are larger than $\frac{\pi}{2}$, i.e., both $\tau^\mathrm{T} (p^1_K(t-~h)-~p^i_\text{target}) < 0$ and
	$\tau^\mathrm{T} \frac{ \partial C^1 }{ \partial p^1_K }|_{p^1_K=p^1_K(t-h)}<0$
	hold.
	In addition, $\tau^\mathrm{T} a^{12}_K = 0$ holds by definition.
	Following~\eqref{eq:LpK}, it can be derived that
	\begin{equation*}
		\tau^\mathrm{T} \frac{ \partial \mathcal{L}^1 }{ \partial p^1_K }|_{p^1_K=p^1_K(t-h)} = \tau^\mathrm{T} \frac{ \partial C^1 }{ \partial p^1_K }|_{p^1_K=p^1_K(t-h)} <0.
	\end{equation*}
	After convex programming, the Lagrange function $\mathcal{L}^1$ will be decreased and $\tau^\mathrm{T}(p^1_K(t)-p^1_K(t-h))>0$ is enforced, i.e., $p^1_K$ has the motion tendency along the direction of $\tau$.

    As shown in Fig.~\ref{figure:unstable_deadlock-2}, both
	${\tau_t}^\mathrm{T} p^1_K \geq {\tau_t}^\mathrm{T} p^1_\text{target}$
	and
	${\tau_t}^\mathrm{T} p^2_K \leq {\tau_t}^\mathrm{T} p^1_\text{target}$ hold at time $t$,
	where $\tau_t$ is the unit vector that is perpendicular to $p^1_\text{target}-p^2_\text{target}$ at time~$t$.
    Due to the motion tendency described above,
    the angle between $p^1_K-p^2_K$ and $p^2_\text{target}-p^1_\text{target}$, denoted as $\beta(t)$,
    is monotonically increased, i.e., $\beta(t+h) > \beta(t)$ when $\beta(t) \neq 0$.
    Note that $\beta(t)$ is bounded by the maximum angle $\beta_\text{max}$, which is the one associated with the common tangent line of these two spheres~$\mathcal{B}^1_d(r_d)$ and~$\mathcal{B}^2_d(r_d)$.
 
    	As already proven, $p^i_{k+1}=p^i_k$ holds for $k \in \mathcal{K}$, which implies $p^i_1 = p^i_K$ for $i=1,2$. 
    	Since $p^i(t+h) = p^i_1$, it is evident that the aforementioned property for $p^i_K$ also holds for $p^i$, $i=1,2$.
    	Consider a situation that robots~$1$ and $2$ are initially centrosymmetric around the point $\frac{p^1_\text{target}+p^2_\text{target}}{2}$ with $p^i(0)=p^i_k(0)$, $k\in\mathcal{K}$, $i=1,2$. Then, it is evident that these two robots will always remain centrosymmetric. 
    Thus, given a bound arbitrarily close to the deadlock equilibrium and $\beta(t_0) $ arbitrarily close to zero, we can find an initial state stated as above for each robot such that robots~$1$ and 2 will leave their corresponding bounds $\mathcal{B}^1_d(r_d)$ and $\mathcal{B}^2_d(r_d)$ simultaneously.
    According to Definition~\ref{def:stable-deadlcok}, the deadlocks for both robots~$1$ and $2$ are unstable.
    This completes the proof.
\end{proof}

\ifCLASSOPTIONcaptionsoff
  \newpage
\fi



%

\bibliographystyle{IEEEtran}
\bibliography{REF}

\end{document}